# The Imitation Game: Detecting Human and AI-Generated Texts in the Era of ChatGPT and BARD

Kadhim Hayawi, Sakib Shahriar, Sujith Samuel Mathew

College of Interdisciplinary Studies, Computational Systems, Zayed university, Abu Dhabi, UAE

## Abstract

The potential of artificial intelligence (AI)-based large language models (LLMs) holds considerable promise in revolutionizing education, research, and practice. However, distinguishing between human-written and AI-generated text has become a significant task. This paper presents a comparative study, introducing a novel dataset of human-written and LLM-generated texts in different genres: essays, stories, poetry, and Python code. We employ several machine learning models to classify the texts. Results demonstrate the efficacy of these models in discerning between human and AI-generated text, despite the dataset's limited sample size. However, the task becomes more challenging when classifying GPT-generated text, particularly in story writing. The results indicate that the models exhibit superior performance in binary classification tasks, such as distinguishing human-generated text from a specific LLM, compared to the more complex multiclass tasks that involve discerning among human-generated and multiple LLMs. Our findings provide insightful implications for AI text detection while our dataset paves the way for future research in this evolving area.

## 1. Introduction

Artificial Intelligence (AI) has progressed rapidly in recent years, contributing to the advent of large language models (LLMs), such as GPT (ChatGPT) [1] and LaMDA (BARD) [2]. LLMs are composed of billions of parameters and built upon pre-existing datasets, leveraging an extensive pre-training process to learn and fine-tune their ability to generate intricate texts. These models are capable of generating human-like text, ranging from essays and stories to software code [3], contributing to a growing digital ecosystem that blurs the boundary between human and machine authorship. Consequently, new questions and ethical challenges have arisen that necessitate careful exploration and research.

The rapid progression of LLMs has led to increasing concerns about their potential misuse. The ability of these models to generate content indistinguishable from that of human authors has implications in a variety of domains, particularly in the field of academic integrity [4], [5], [6]. The emerging challenge faced by educators and institutions today is the potential misuse of these sophisticated models by students for generating academic essays or coding assignments, threatening the core principles of academic integrity.

Hence, there is a pressing need for effective methods to differentiate between human-generated and machine-generated text.

To address these concerns, the first objective of this research is to introduce a novel dataset comprising human and LLM-generated texts, spanning a wide range of genres, including essays, stories, poetry, and software code. The uniqueness of this dataset lies in its broad coverage of text types and its dual-source nature, making it a valuable resource for exploring the nuances between human and machine writing. Our second research objective focuses on the comparative analysis of machine learning models to classify texts based on their source – human or LLM. By investigating various models' performance, we aim to understand the strengths and weaknesses of existing techniques in identifying LLM-generated responses. Our findings will not only serve to inform the development of more robust AI models but also assist in the creation of academic integrity checks in an increasingly digitized educational environment.

The rest of this survey is organized as follows. Section 2 presents the necessary background information. Section 3 describes the creation and composition of our dataset. Section 4 presents the methodology and results of our comparative study of ML models, including model selection and performance comparison. Section 5 discusses the implications of our results and hints at future research directions, while Section 6 concludes the paper.

## 2. Background

In this section, we present a concise background on large language models and their impact on academic integrity. We also discuss machine learning and text processing concepts used in this work.

### *2.1 The Impacts of LLMs on Academic Integrity*

The remarkable progress in Artificial Intelligence (AI) over the last several decades has profoundly reshaped numerous aspects of modern life, impacting various fields from healthcare to entertainment [7], [8]. At the forefront of this revolution are Language Learning Models (LLMs), a subset of AI that has fundamentally transformed language generation. The evolution of LLMs traces back to the development of Machine Learning (ML) algorithms that could be trained to predict future data points based on the patterns identified in the original data. However, the transformative shift in the field came with deep learning, a class of ML models designed to automatically and adaptively learn intricate structures from high-dimensional data. These advancements culminated in the creation of sophisticated LLMs like GPT [1], capable of generating human-like text that's nearly indistinguishable from the work of a human author. LLMs leverage deep learning, human feedback, and vast datasets to understand and generate human-like text. These models, trained on numerous text inputs, can predict subsequent words in a sequence, thus generating coherent and contextually relevant sentences. GPT-3, with its 175 billion parameters, can perform a variety of language tasks, such as translation, question answering, and text generation, without specific task-oriented training [3]. However, what sets LLMs apart is their capability to generate creative content, like stories or essays, which is a complex task that requires a nuanced understanding of language. Their capability to grasp the style, tone, and contextual subtleties of language makes them powerful tools for numerous applications, ranging from content creation to automated customer service [9]. Despite their impressive abilities, LLMs aren't without limitations. They can occasionally generate nonsensical or inappropriate content and are highly sensitive to input phrasing [10], [11], [12]. These limitations open up areas of research to further enhance the capabilities and safety of LLMs [13], driving the pursuit of more sophisticated AI models.

Despite these significant advancements, differentiating between human-generated and AI-generated text presents a unique set of challenges. The sophistication of current LLMs means they can mimic human-like stylistic nuances, making their outputs more challenging to distinguish [14]. Ethical considerations also emerge as the misuse of these AI models to generate misleading or harmful content becomes a growing concern [15]. Furthermore, the real-world implications of these challenges are significant, with potential ramifications in fields like journalism, online content creation, and academic integrity. Amid escalating concerns over academic integrity, the sophistication and accessibility of LLM-based text generators necessitate the ability to distinguish between human and AI-generated text, as the potential misuse of these technologies threatens to undermine the educational system's foundational principles of originality and individual effort. The advent of LLMs poses substantial challenges to maintaining academic integrity. Firstly, their ability to generate high-quality, contextually relevant text makes it possible for students to use such technology to complete assignments or papers, instead of writing original content [16]. This could lead to a concerning rise in plagiarism and intellectual dishonesty by submitting work that is not the product of the individual's understanding or effort. Moreover, the broad and nuanced knowledge base that these LLMs possess can potentially outperform human capabilities in various academic tasks. For instance, a student could leverage an AI text generator to write an essay on a topic they barely understand, thus bypassing the learning process entirely. This fundamentally disrupts the purpose of education, which is to develop a person's knowledge, understanding, and critical thinking skills. Finally, the sophistication of these models could also compromise the detection methods that educational institutions employ to identify academic misconduct. Traditional plagiarism detection tools may not identify AI-generated text as plagiarized [17] since it generates unique content based on human-written patterns. Thus, while LLMs offer exciting possibilities for many applications, they also threat academic integrity [4], [5]. Educators, institutions, and AI developers must understand and address these challenges, developing new strategies and tools for preserving the principles of honesty and originality in the academic world.

## *2.2 Machine Learning and Deep Learning*

ML aims to instruct an algorithm or system the ability to learn from and improve upon its performance through exposure to data, without the requirement of explicit programming. This iterative learning process harnesses various algorithms to find patterns, thus enabling the system to discover insights and make predictions independent of human intervention. We've employed three widely recognized traditional ML algorithms in our work:

1. Random Forest (RF): RF is an ensemble learning method, generating numerous decision trees and amassing their results for the final prediction [18]. It capitalizes on the diversity of the decision trees it spawns, each crafted from different subsets of the original dataset. This collective intelligence helps in reducing overfitting, as the ensemble of weak, yet collectively strong decision trees influence the predictions. Moreover, RF also offers insights into the most influential features guiding the model's decisions through feature importance metrics.
2. Support Vector Machine (SVM): SVMs [19] seek the hyperplane in the feature space that maximizes the margin between the two classes. The optimal hyperplane ensures maximum distance from the nearest training data points across all classes. SVMs are robust against overfitting and efficient in high-dimensional spaces, even when the number of dimensions exceeds the samples. They can manage non-linear boundaries, depending on the kernel function used. We opted for a linear kernel, which implies a straight-line (or hyperplane) decision boundary.
3. Logistic Regression (LR): LR is a binary classification model that calculates the probability of an instance belonging to the default class [20]. LR's output transforms a logistic (or sigmoid) function, confining the output between 0 and 1, thus denoting a probability. Despite its seeming simplicity, LR's interpretability and ability to handle non-linear effects make it a commonly used baseline in machine learning studies.

Complementing traditional ML, we also utilize deep learning (DL) in our study. DL algorithms contain artificial neural networks with multiple hidden layers to model and understand complex patterns in datasets. Its strength lies in processing unstructured data and tackling problems with high dimensionality, making it ideal for various tasks, including natural language processing, image recognition, and speech recognition [21], [22]. Long Short-Term Memory (LSTM) networks, the DL model utilized in our study, are a special type of Recurrent Neural Networks (RNNs) capable of learning long-term dependencies. Traditional RNNs encounter difficulties in learning these long-range temporal dependencies due to the vanishing gradient problem, where the importance of input decreases exponentially over time. LSTM networks overcome this issue with a unique design incorporating memory cells, input gates, forget gates, and output gates [23]. These elements complement each other to regulate the flow of information, selectively remembering crucial data over longer periods and discarding non-essential data, thereby making LSTMs a powerful tool for sequential data like text. Its ability to remember patterns over long sequences makes it suitable for differentiating between human and AI-generated text.

### *2.3 Text Processing*

Text processing, a crucial stage in any Natural Language Processing (NLP) task, involves transforming raw text data into a format that ML algorithms can understand and analyze. This process encompasses several steps, including feature extraction, tokenization, vectorization, stop word removal, and stemming, which are fundamental to preparing the text data for ML models.

Feature extraction is the initial step in text processing where specific characteristics or attributes are identified from the raw text data for predictive modeling. In NLP, these features often relate to the words, characters, or groups of words within the text. Tokenization, a crucial part of feature extraction, breaks down the text into individual words or tokens. After tokenization, the next step is vectorization, where text is converted into vector representation for training the ML models. Methods like a bag of words, term frequency-inverse document frequency [24], and word embeddings like GloVe [25] transform the textual data into numerical vectors. In many texts, certain words are frequently used but provide little to no useful information. These stop words include articles, prepositions, and pronouns, and removing them can help reduce the dimensionality of the data and increase computational efficiency [26]. Stemming reduces inflected (or sometimes derived) words to their root form, i.e., base or stem. For instance, stemming should reduce the words "playing", "plays", and "played" to the base word, "play." This helps in consolidating similar features into one, reducing the overall complexity.

## 3. Dataset Collection

The collection of a diverse and comprehensive dataset plays a crucial role in training and evaluating ML models. Particularly for our research, which aims to classify texts generated by LLMs and human authors, having a representative dataset is significant. Firstly, by aggregating responses from various LLMs, our dataset covers a comprehensive spectrum of their textual generation capabilities given that different models employ unique strategies. For this research, we have gathered responses from the popular GPT (ChatGPT) and LaMBDA (BARD) models. To the best of our knowledge, this is the first dataset offering a broad perspective on the state-of-the-art AI language generation. Furthermore, integrating human-generated responses into our dataset is essential as it provides a valuable comparison point and allows us to benchmark the performance of these AI models against human authors. Another aspect of our data collection process was the inclusion of different categories of texts; essays, stories, poems, and software code. These categories represent a wide range of contexts and styles a model might encounter in real-world applications. In total, we gathered responses from GPT, BARD, and human authors across several of these categories. This approach ensures our dataset's diversity, offering a comprehensive view of current language generation abilities. Finally, to contribute to the academic community, we are making our dataset publicly available for researchers. We believe that access to such a dataset will aid researchers in the development of more accurate models for distinguishing between LLM-generated and human-generated texts, propelling the field of academic integrity in the AI era forward. The dataset can be accessed from GitHub (https://github.com/sakibsh/LLM).

### *3.1 Essays Dataset*

A portion of our human essay dataset was collected from the Automated Student Assessment Prize, sponsored by the William and Flora Hewlett Foundation [27]. The foundation's initiative aimed at appealing to data scientists and machine learning specialists to develop quick, efficient, and affordable solutions for automated grading of student-written essays. Alongside this, we sourced data from the "essays with instructions" dataset available on Hugging Face. This dataset, created and curated by Christoph Schuhmann, offers a vast and diverse range of student essays, each accompanied by specific writing instructions [28]. Furthermore, we also incorporated a dataset from the research work on the BB-SVM model for automatic personality detection [29]. This dataset, labeled with big-five personality traits, not only enriched the diversity of our collection but also introduced an element of psychological analysis to our research. In total, we collected 17,684 human-written essays. This expansive dataset provides a solid foundation for our ML model training, helping us understand the nuances and intricacies of human language use and facilitating the ML model to differentiate between human-generated and AI-generated essays.

To collect responses from GPT-3.5, we leveraged 200 prompts from the book *501 Writing Prompts* by Learning Express (2014) [30]. These prompts, encompassing a broad range of topics and themes, were used to solicit responses from GPT-3.5. As a result, we were able to amass a rich collection of AI-generated essays that showcased the capacity of GPT-3.5 to generate human-like text across diverse topics. Similarly, we used the same set of 200 prompts to obtain responses from BARD. This approach allowed us to maintain consistency in our dataset collection process and enabled adequate comparison between the capabilities of GPT-3.5 and BARD. Figures 1 and 2 display the comparison of word cloud visualization and histogram for the human-generated, GPT, and BARD essay datasets, respectively.

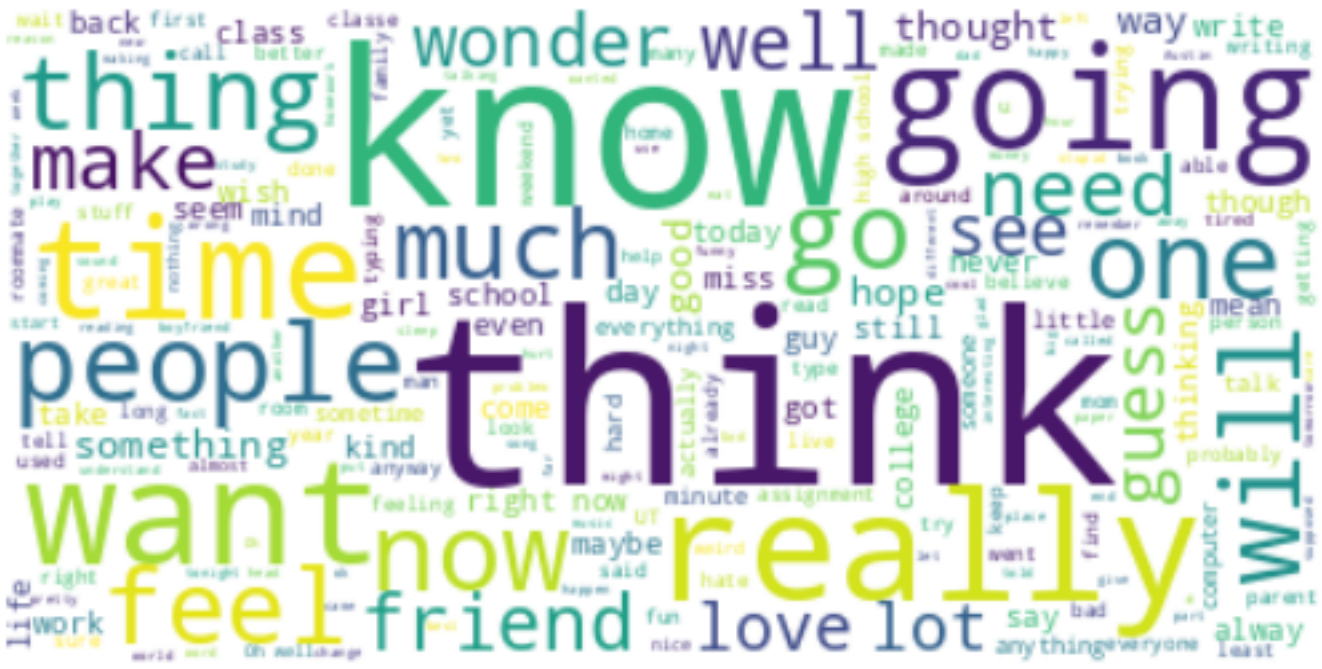

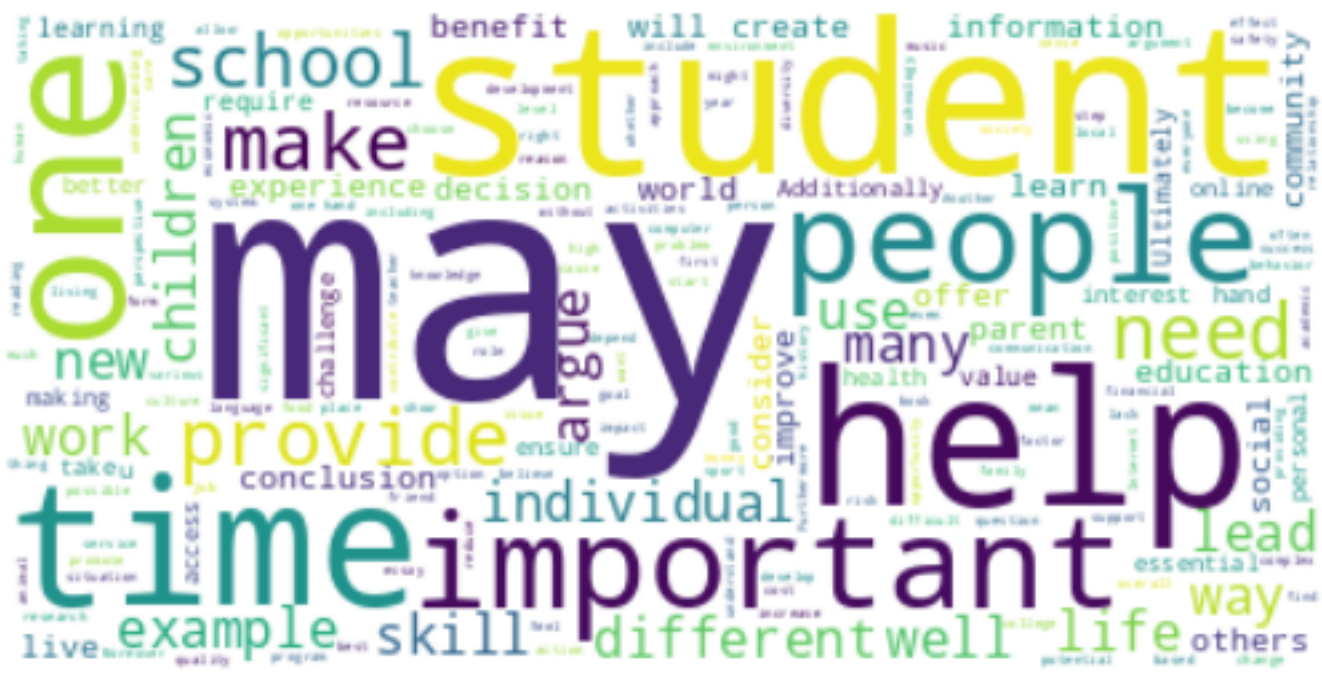

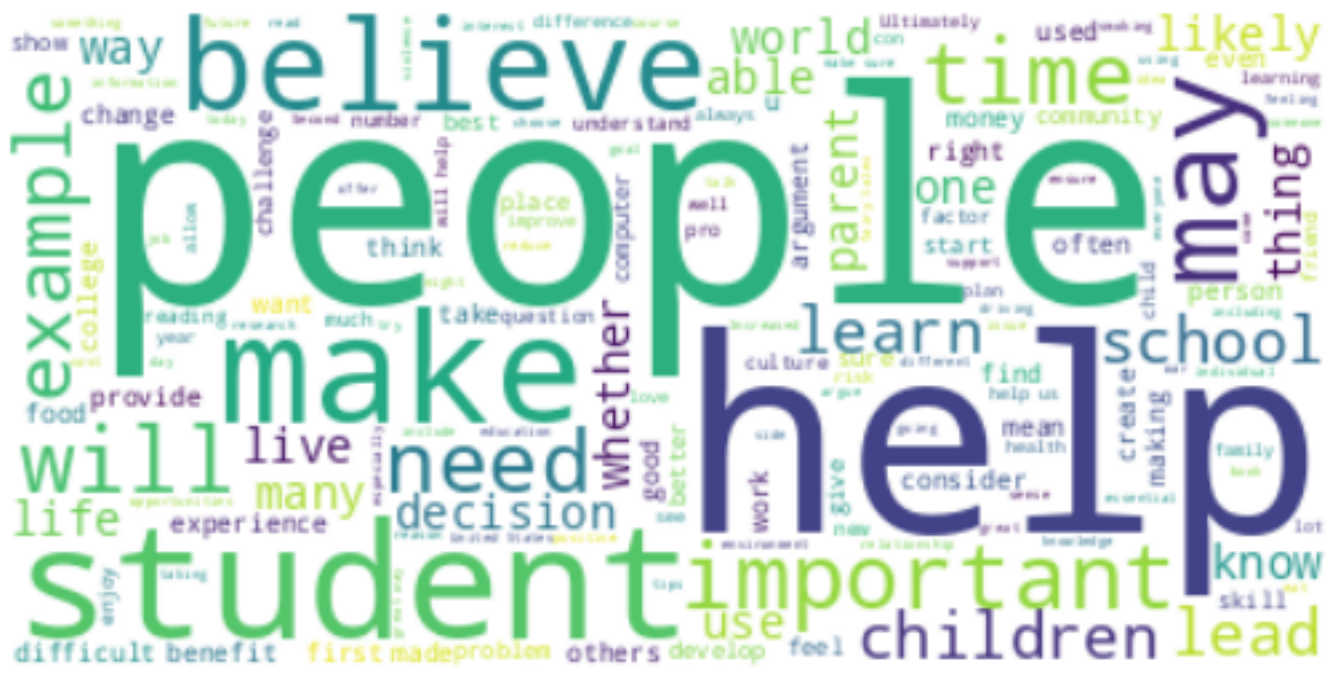

**Figure 1 Word Cloud Visualization for Human-generated (top), GPT (middle), and BARD (bottom) essays**

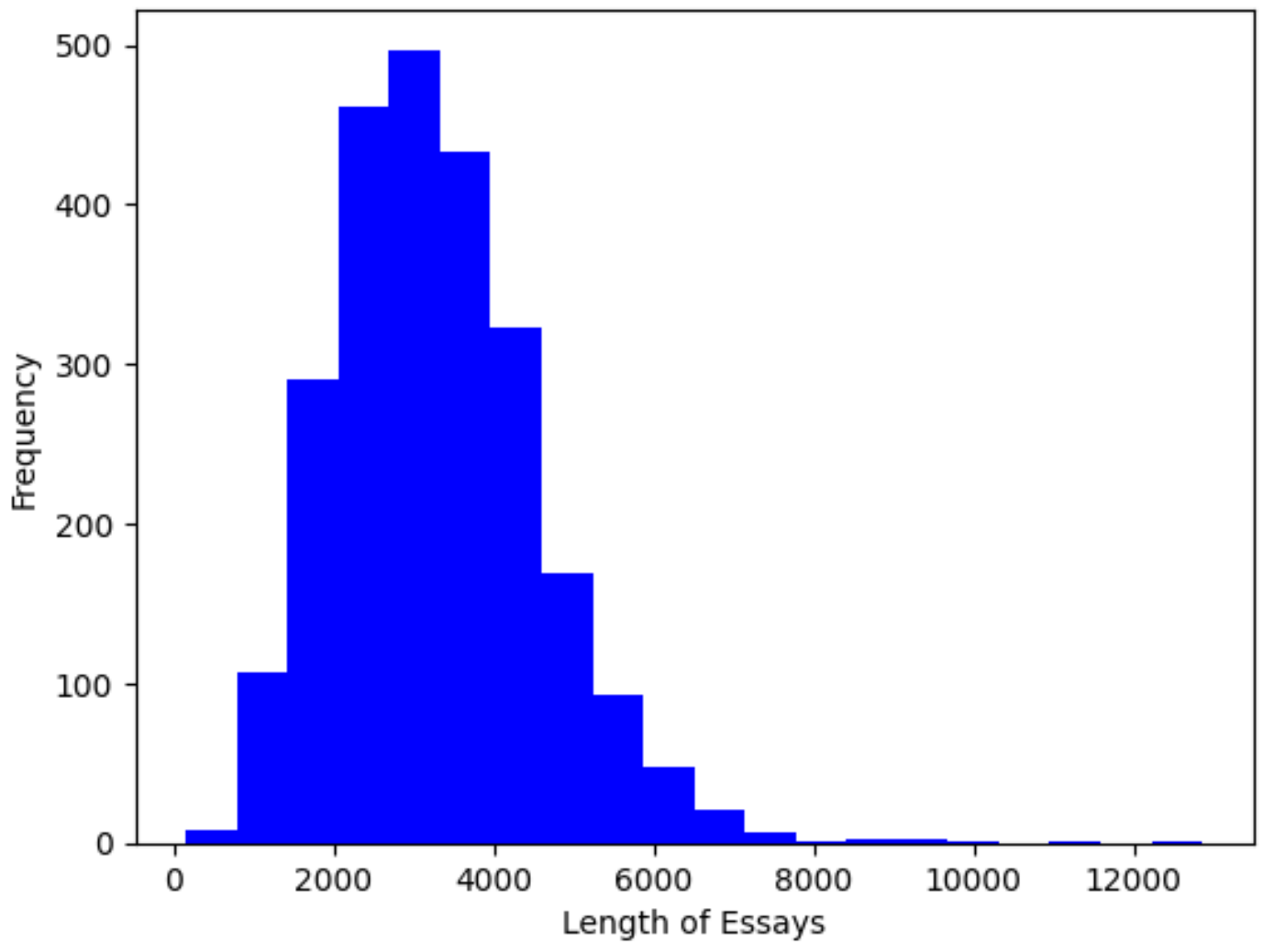

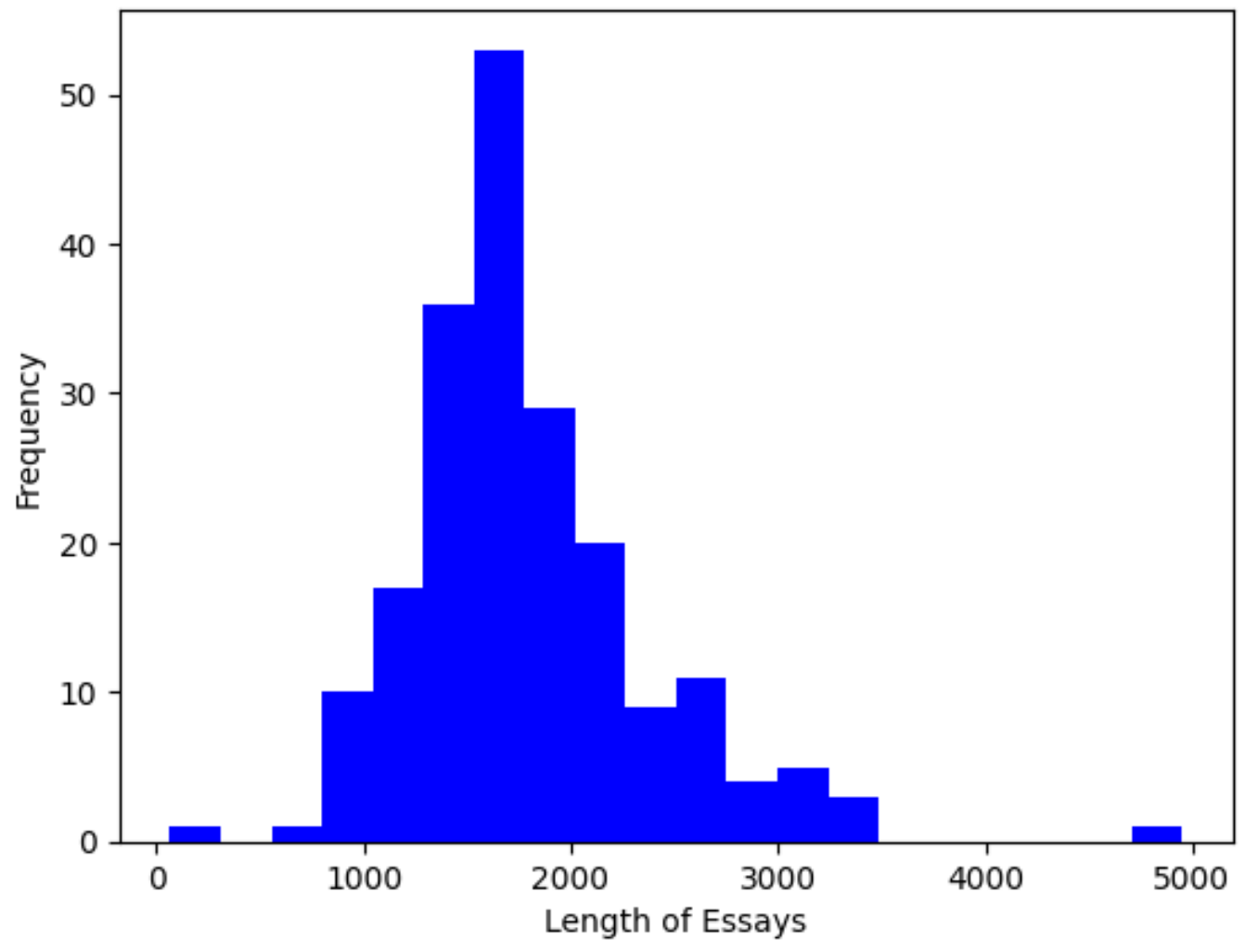

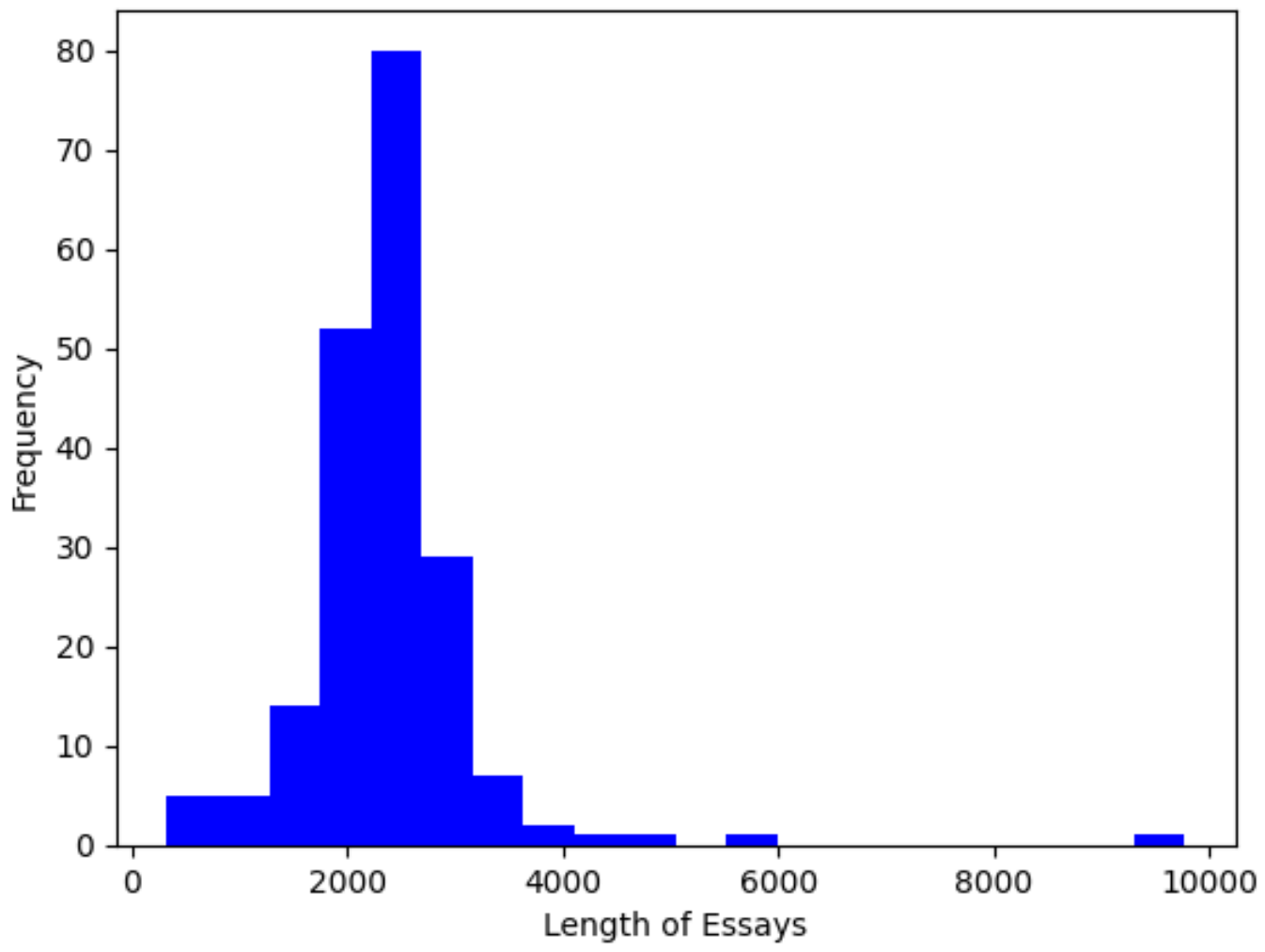

**Figure 2 Histogram for Human-generated (top), GPT (middle), and BARD (bottom) essays**

### 3.2 *Stories Dataset*

To collect stories or fiction authored by humans, we leveraged the comprehensive database of the "Archive of Our Own" [31], an initiative by the Organization for Transformative Works (OTW). Established in 2007, OTW is a non-profit entity devoted to the preservation and accessibility of fanworks and fan culture. The *Archive of Our Own*, one of OTW's key projects, is a non-commercial platform built on open-source archiving software that hosts an extensive array of fanworks.

In our data collection process, stories tagged with 'ChatGPT' were classified as GPT, and the remaining stories were considered as human-written. We developed and implemented a Python script to automate the data collection, enabling efficient parsing of the platform to extract the relevant stories for our dataset. The final count yielded 180 human stories and 95 GPT-3 generated stories. Since we did not find any BARD-generated stories in this archive, we obtained BARD responses through prompts. We used GPT-3 to generate 300 unique story-writing prompts and obtained responses from BARD to these prompts. Figures 3 and 4 display the comparison of word cloud visualization and histogram for the human-generated, GPT, and BARD story datasets, respectively.

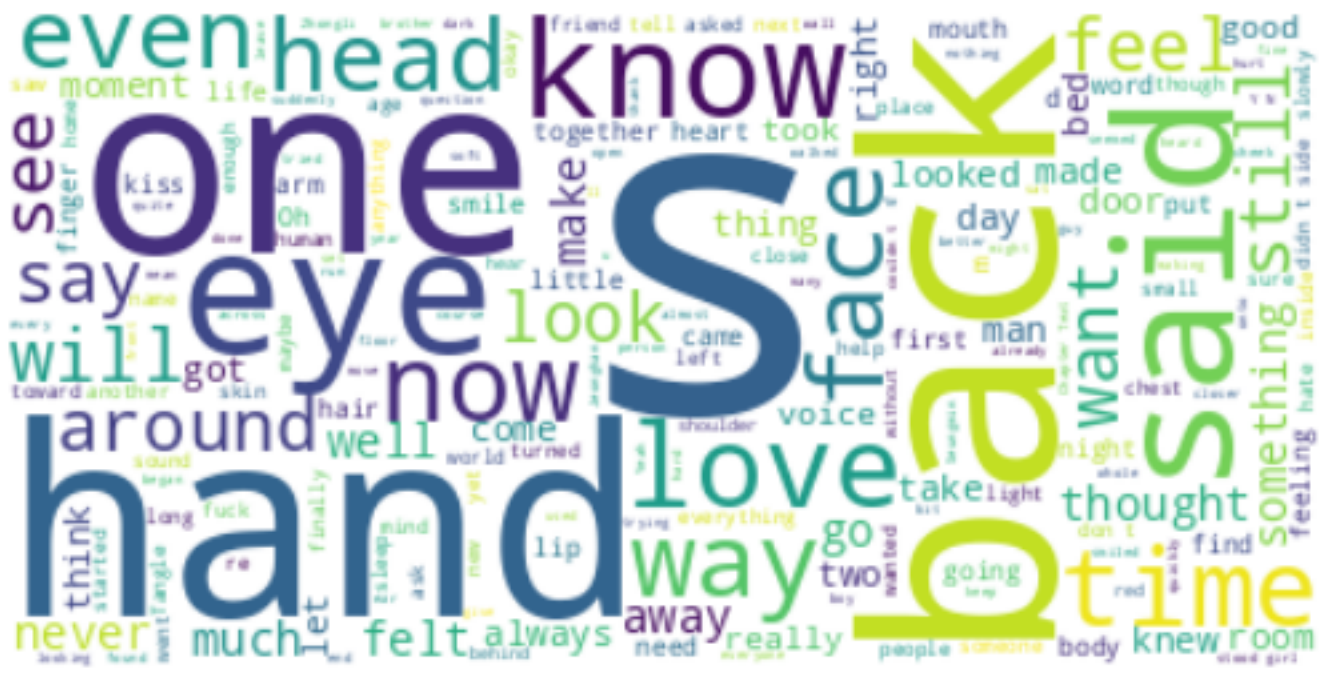

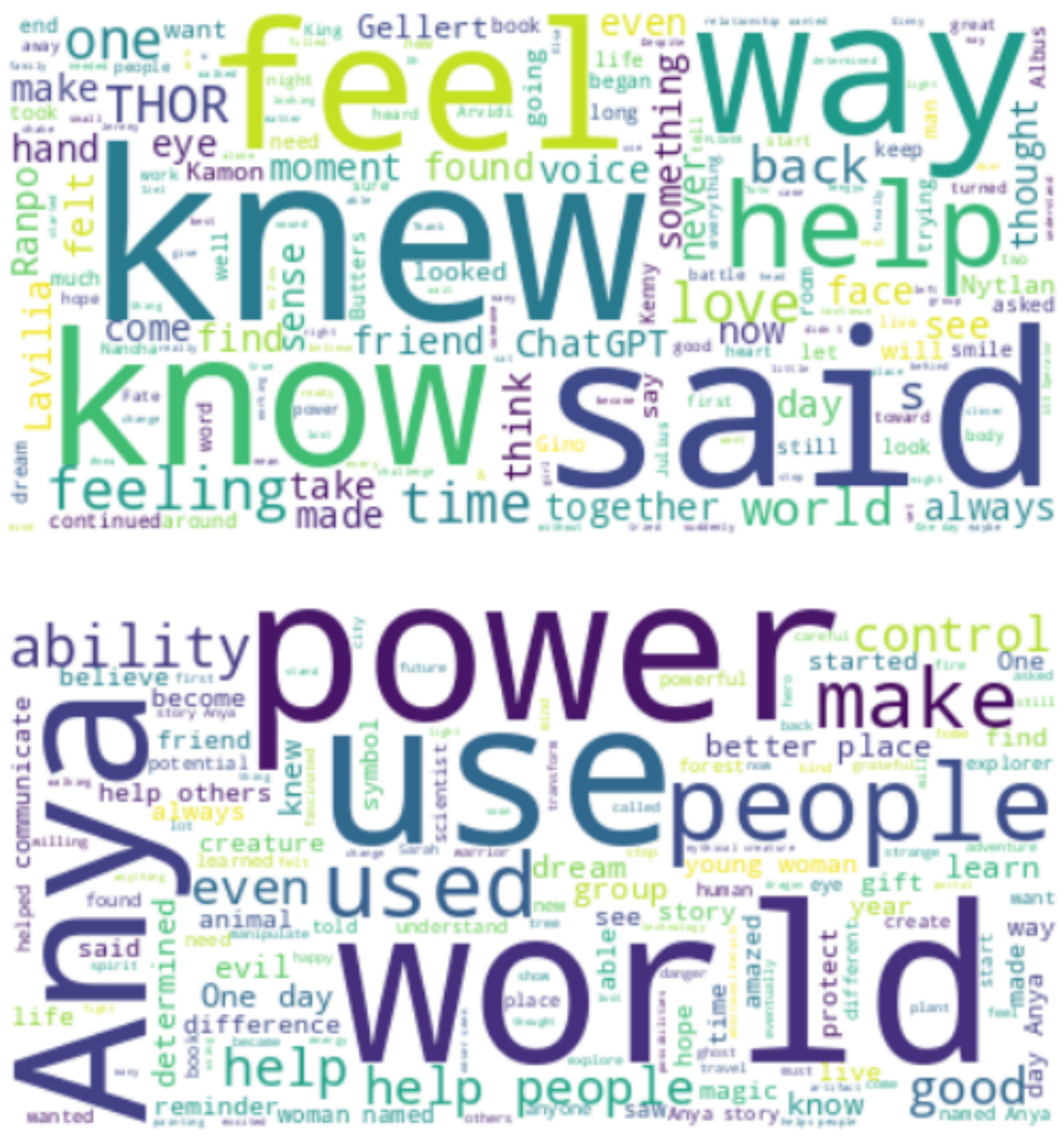

Figure 3 Word Cloud Visualization for Human-generated (top), GPT (middle), and BARD (bottom) stories

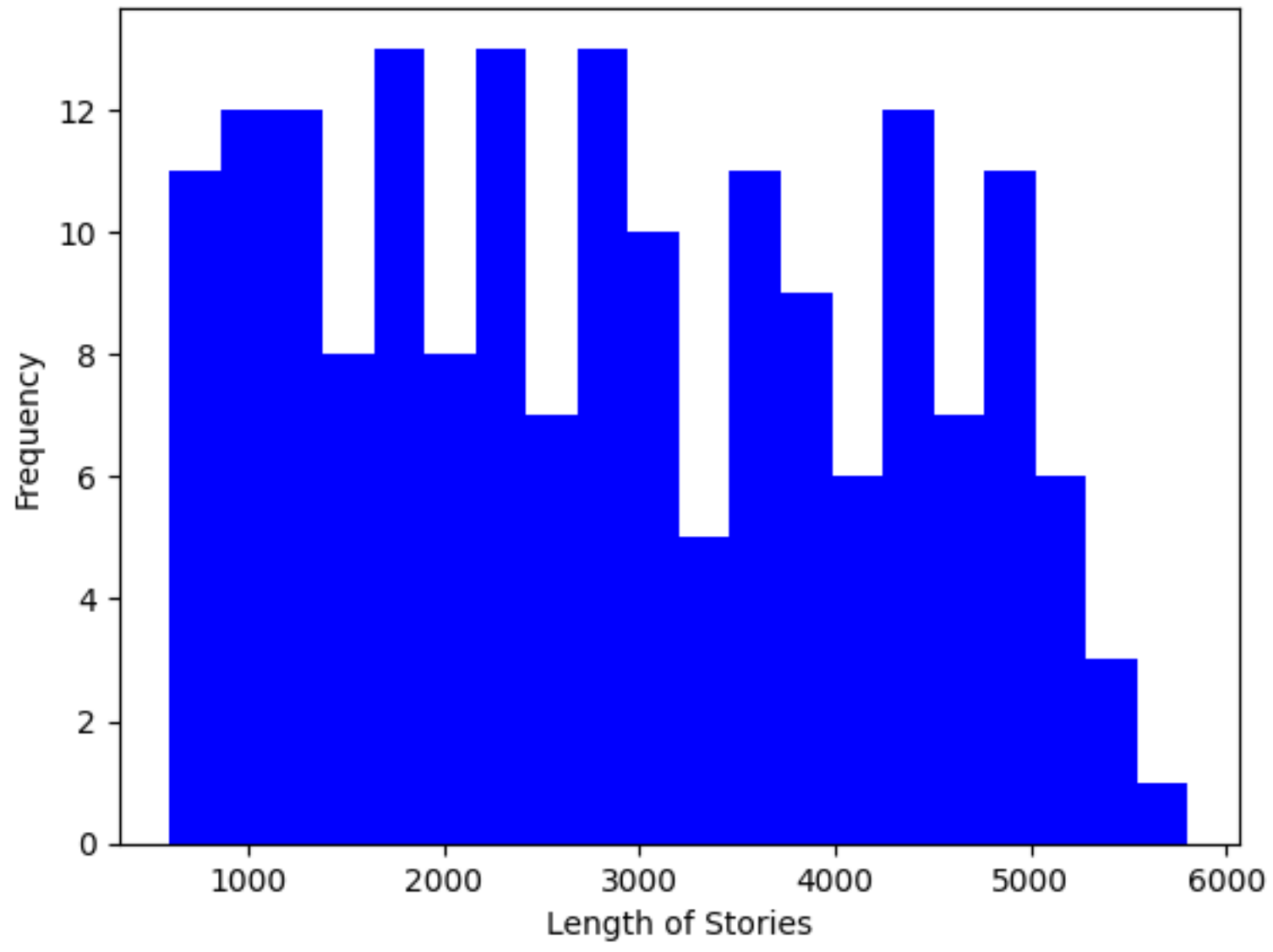

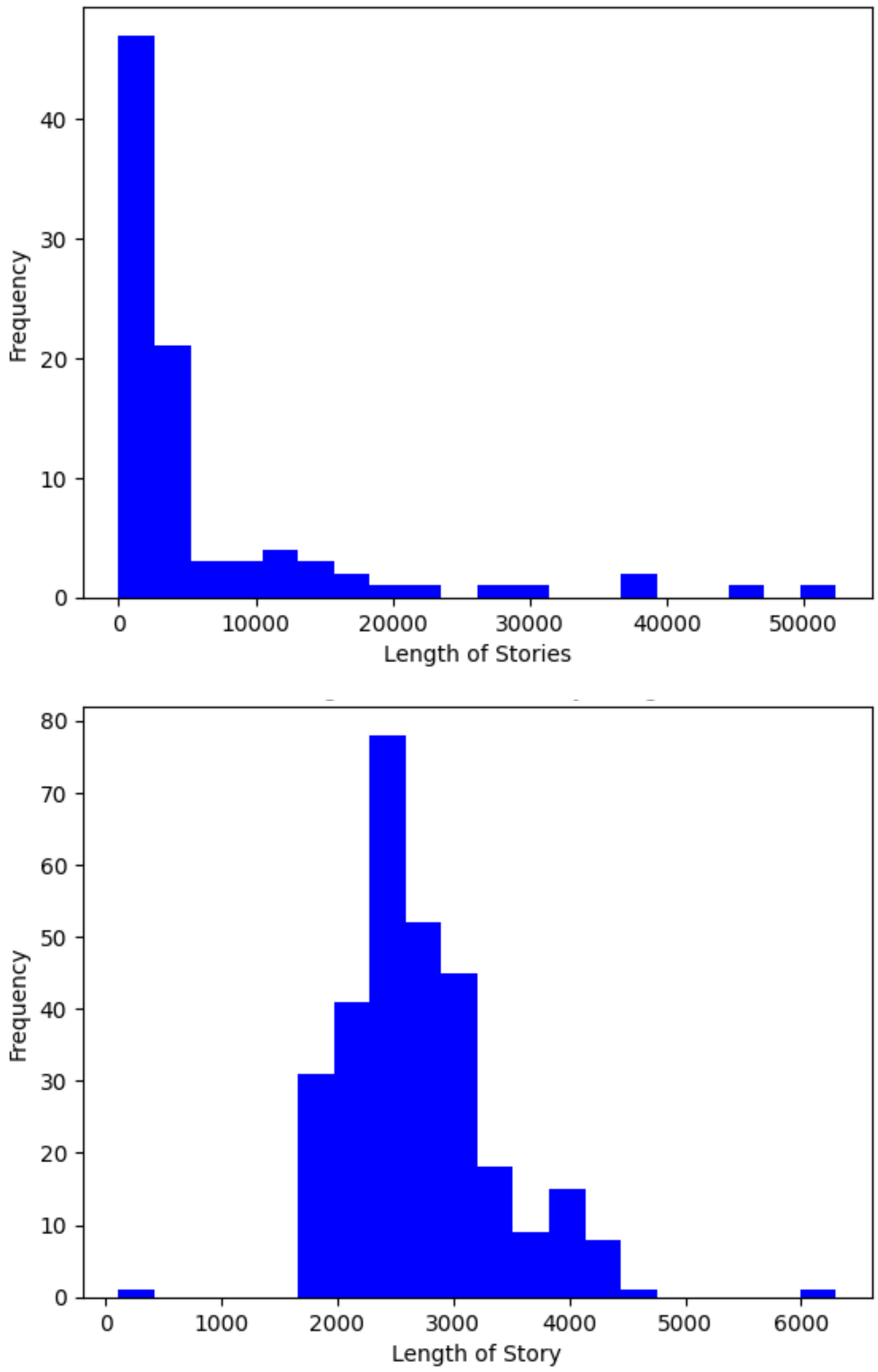

**Figure 4 Histogram for Human-generated (top), GPT (middle), and BARD (bottom) stories**

### *3.3 Poetry Dataset*

For human-written poems, we used the dataset provided by *PoetryFoundation.org*, a literary organization that emerged in 2003 after receiving an endowment from philanthropist Ruth Lilly [32]. The Foundation focuses on augmenting the role and reach of poetry, striving to sustain environments for everyone to create, experience, and share poetry. The collected dataset is a comprehensive collection, encompassing a total of 13,854 human-written poems.

To collect poems written by GPT, we utilized several websites like *thinkwritten.com* [33], which provided an assortment of poetry writing prompts. Using these prompts, we elicited a total of 250 poems written by GPT-3.5. To compare across LLMs, we used the same prompts for BARD and collected a corresponding total of 250 poems. Figures 5 and 6 display the word cloud visualization and histogram for the human-written, GPT, and BARD poetry datasets, respectively.

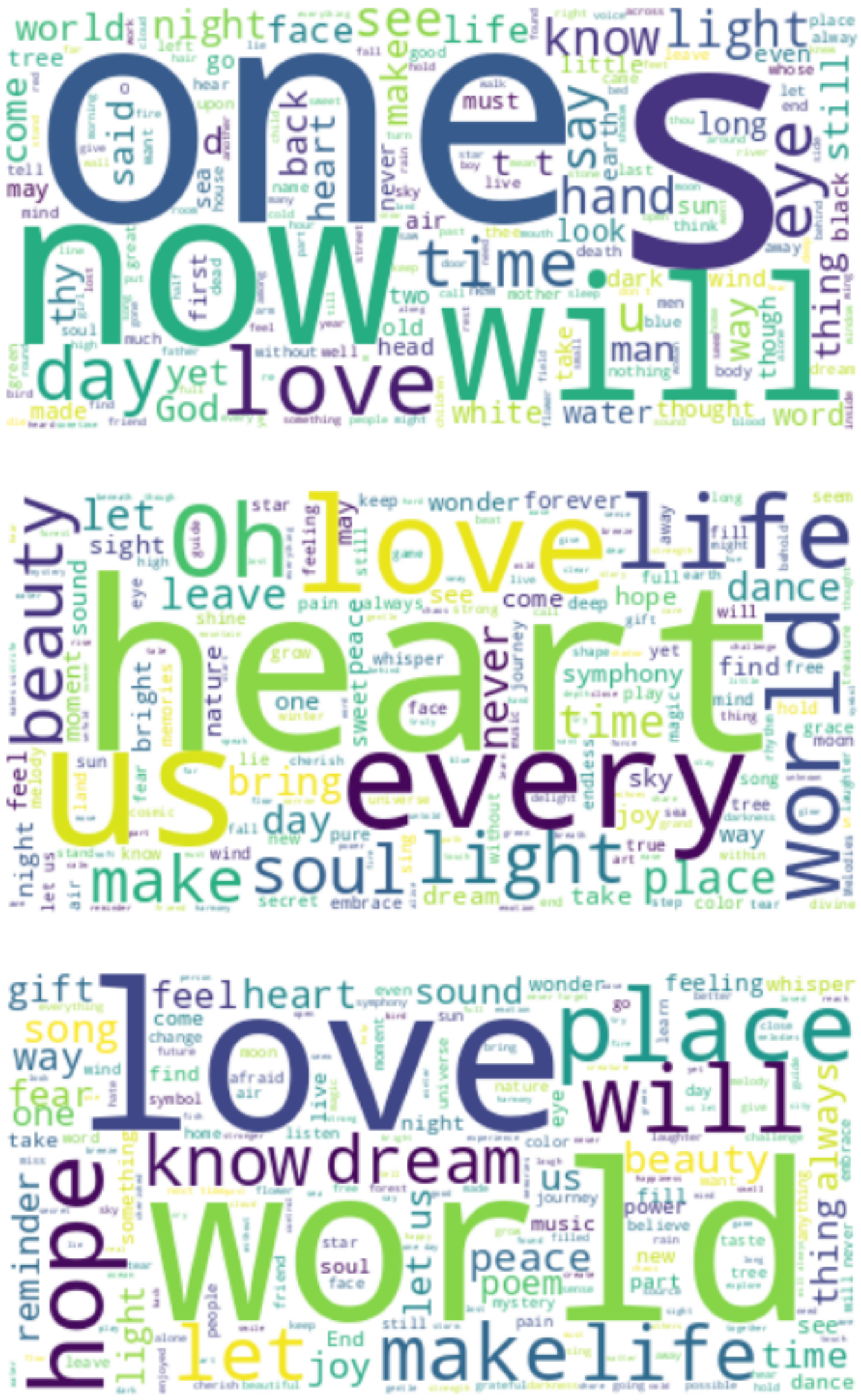

**Figure 5 Word Cloud Visualization for Human-generated (top), GPT (middle), and BARD (bottom) poems**

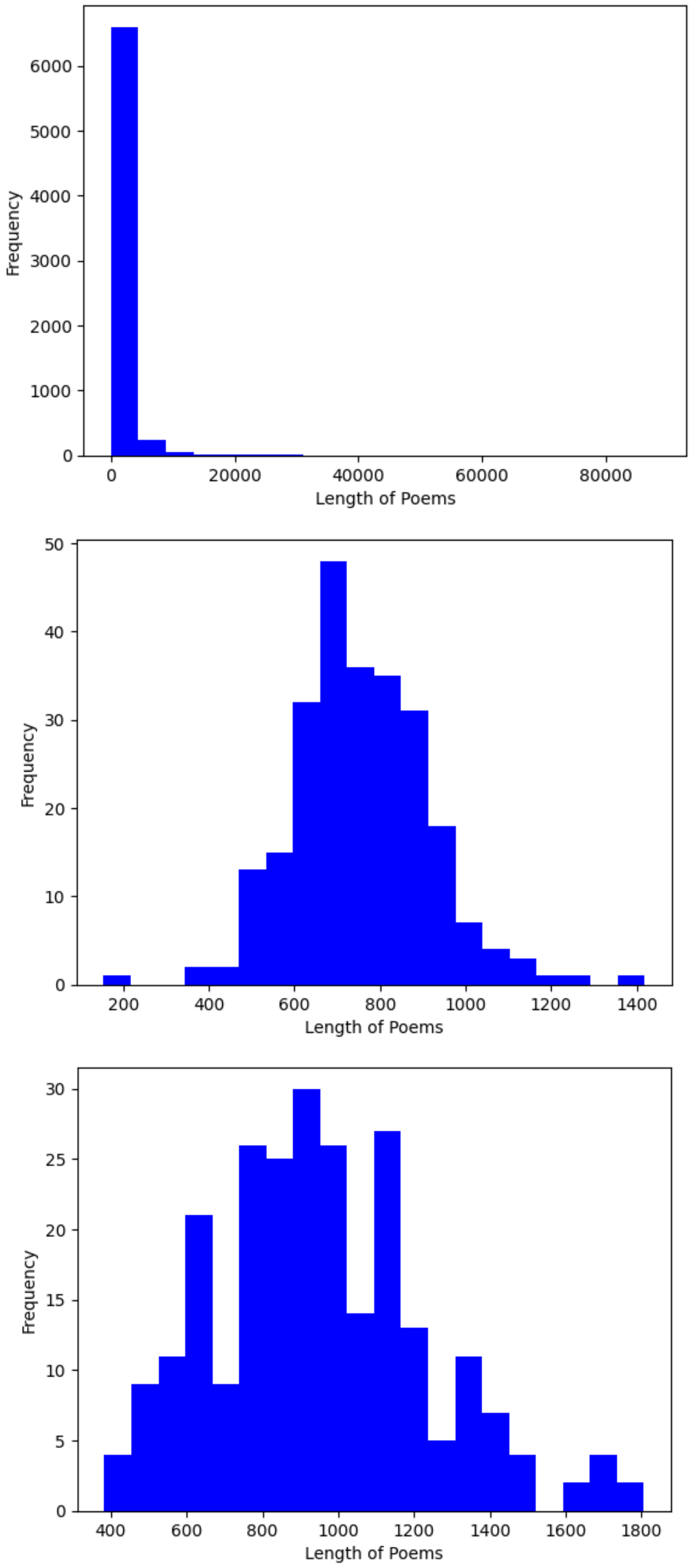

**Figure 6 Histogram for Human-generated (top), GPT (middle), and BARD (bottom) poems**

### 3.4 *Python Code Dataset*

In this study, we gathered Python code generated by GPT and BARD. This approach entailed procuring code writing prompts available online, resulting in a collection of 249 code responses from each of these LLMs. For the code-generation prompts, we relied on commonly encountered problems in programming and software development, such as sorting algorithms, basic arithmetic operations, and data structure manipulation. This ensured the responses would be representative of a broad range of coding tasks.

Our current dataset does not contain any human-written Python code. This is a limitation that we intend to address in future work by including a comparable number of human-authored codes. Potential sources for this could be platforms like GitHub or existing datasets such as the 150k Python Dataset by Raychev *et al*. [34]. It is worth mentioning that the latter primarily contains Python 2.7 code, which may not be suitable for comparison with the Python 3 code produced by the LLMs. Figures 7 and 8 highlight the word cloud visualization and histogram for the GPT and BARD Python code datasets, respectively.

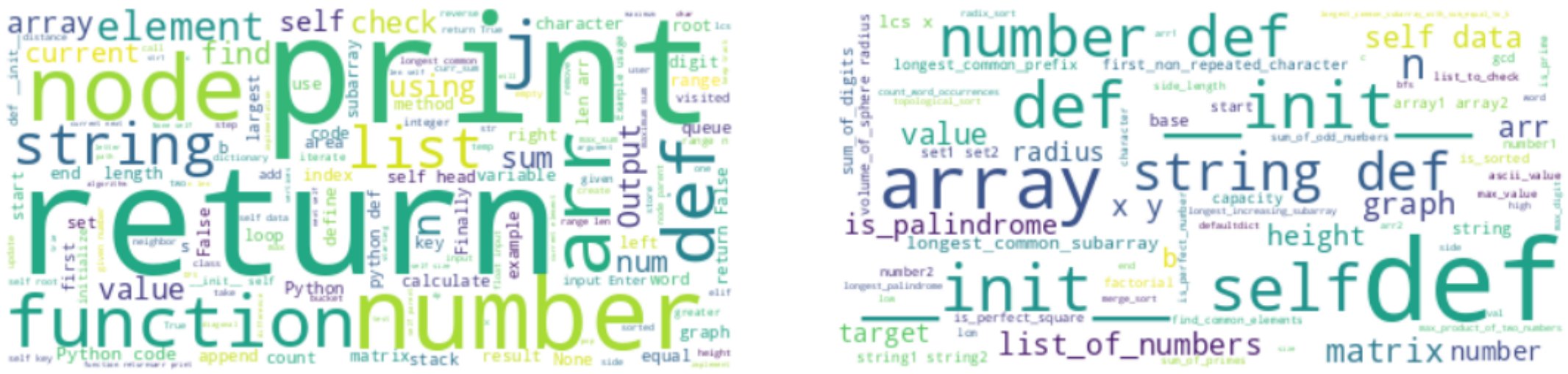

**Figure 7 Word Cloud Visualization for GPT (left) and BARD (right) python code**

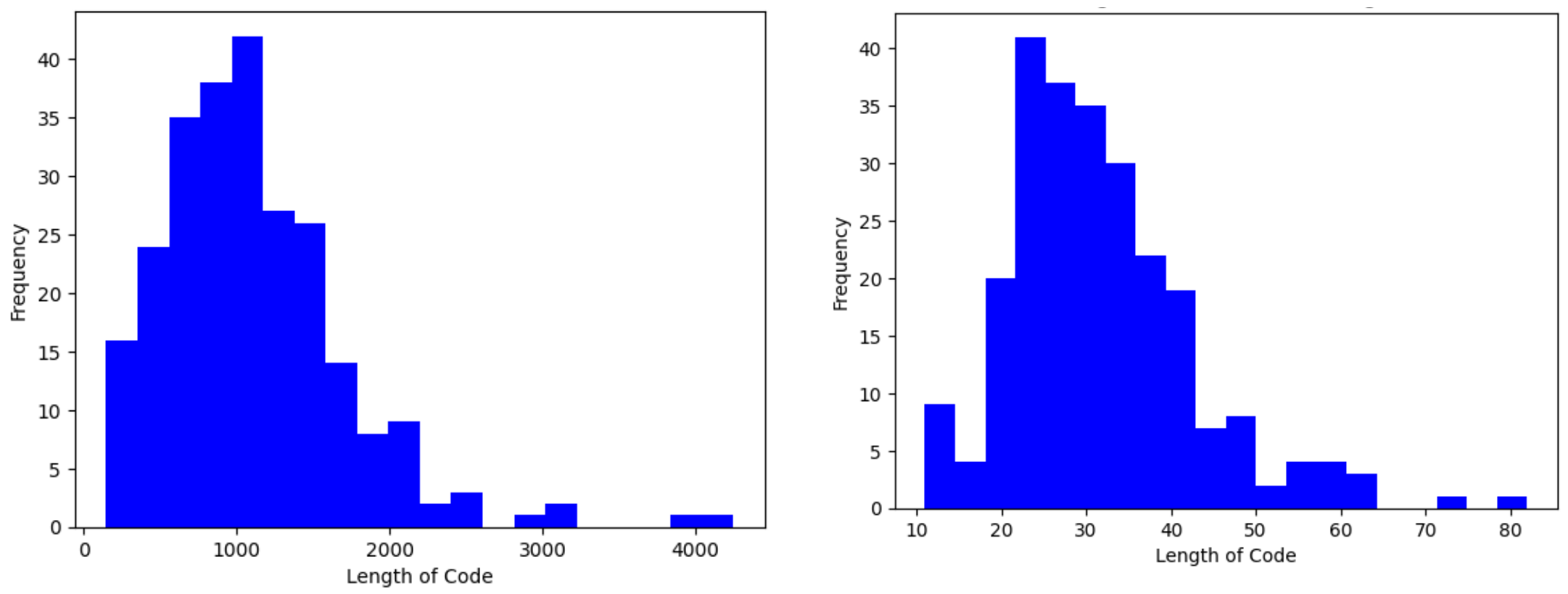

**Figure 8 Histogram for GPT (left) and BARD (right) python code**

## 4. Classification and Evaluation of Human-Generated and LLM texts

In this section, we outline the experimental setup for the classification of human-generated texts and LLM texts. We briefly discuss the text processing and model development steps and also present the classification results from our experiments.

### *4.1 Experimental Setup*

Our experimental setup was designed to evaluate the ability of different ML models to classify essays, stories, and poems generated by humans, GPT, and BARD. For each of these datasets, we conducted two types of experiments. The first experiment involved a multiclass classification task. Here, the ML models were trained to classify a given text as either human-written, GPT-generated, or BARD-generated. In the second experiment, we reduced the complexity of the task by performing binary classification. Specifically, we developed the models to classify human-written texts against GPT-generated (human vs GPT) and human-written texts against BARD-generated (human vs BARD). Across all these experiments, we maintained a consistent approach for the ML models employed. We utilized four models widely used in text classification tasks: Random Forest (RF), Support Vector Machine (SVM), Logistic Regression (LR), and Long Short-Term Memory (LSTM) networks. The transformation of each text into numerical features was accomplished through a pipeline that integrated CountVectorizer and TfidfTransformer. The CountVectorizer converts each text into a "bag of words" structure and the TfidfTransformer assigns weights to these counts to quantify the importance of individual words.

The LSTM model starts with an embedding layer that converts each input word into a dense vector of fixed size, capturing the semantic meaning of the word. This is followed by a Bidirectional LSTM layer with 50 units that can understand the context from both past and future time steps. Next, a dense layer with a sigmoid activation function connects all neurons in the LSTM layer, providing a probability output for binary classification problems. The model is compiled using the Adam optimizer [35] and *binary_crossentropy* as the loss function. For a multiclass problem, the loss function would be *categorical_crossentropy* to accommodate multiple classes. Classification performances were evaluated using the following metrics (Equations 1-4). To get a comprehensive evaluation, we used the k-fold cross-validation technique [36] and presented results aggregated across 5-folds.

$$Accuracy = \frac{TP + TN}{TP + TN + FP + FN} \tag{1}$$

$$Recall\ (TPR) = \frac{TP}{FN + TP} \tag{2}$$

$$Precision = \frac{TP}{TP + FP} \tag{3}$$

$$F1\,Score = \frac{2 * Precision * TPR}{Precision + TPR} \tag{4}$$

## *4.2 Data Cleaning and NLP*

Given the heterogeneous nature of our dataset, standardizing and cleaning the data was a crucial step in training the ML models. We first harmonized the data frames of each source by adding a 'source' column to indicate the origin of each text and then concatenated the data frames. This led to a unified dataset containing texts from all three sources, further refined by removing unnecessary columns, such as *ID* and *prompts*. Next, we performed text-cleaning operations to ensure the quality of our data. We used regular expressions to remove URLs, HTML-encoded characters, and any extra whitespaces from the texts. All texts were converted to lowercase and all punctuation except apostrophes was removed.

In preparing our data for ML algorithms, we employed several NLP techniques. We removed common stop words, which often carry little semantic weight, to focus on the words that contribute the most to each text's meaning. We also supplemented the stop words list with additional words ('u', and 'c') observed to be frequent but less informative in our context. To reduce the complexity of the data, we applied stemming using the Snowball Stemmer [37]. This process transforms words to their base or root form (e.g., 'running' to 'run'). Lastly, we encoded the 'source' column to numerical labels, effective for the ML algorithms.

## *4.3 Feature Extraction and Model Development*

We employed a pipeline-based approach for feature extraction and model building, using both traditional ML models and a DL model. The first stage utilized CountVectorizer to convert the essays into a matrix of token counts, transforming our textual data into a numerical form suitable for further processing. Subsequently, we used TfidfTransformer to normalize these counts and reduce the impact of frequent words that may not be as meaningful as rarer words. Finally, we applied the respective classifier to perform the classification task. We evaluated each model using stratified 5-fold cross-validation to maintain the proportion of each source in each fold and obtain a more generalizable estimate of the model's performance. During each fold, we obtained predictions and measured the accuracy of these predictions, repeating this process for all folds. Finally, we averaged the accuracy scores across all folds and calculated their standard deviation.

After completing this process for all the models, we then moved to our DL model. Before developing the LSTM model, we tokenized the cleaned essays into sequences of integers. We then calculated the maximum sequence length and the vocabulary size from the tokenized data. The LSTM model consisted of an embedding layer that mapped our vocabulary to a smaller dimensional space, a bidirectional LSTM layer, and a dense output layer with either three neurons for multiclass classification or two neurons for binary classification. We trained this model using 5-fold cross-validation, similar to the previous models. During each fold, we trained the model on the training data and evaluated its performance on the validation data. After all folds were completed, we calculated the average accuracy and loss across all folds and their standard deviations. We also generated a learning curve to visualize the model's training process and ensure no significant overfitting of the model.

## 4.4 Results on Multiclass classification

In this section, we present the results for multiclass classification using several ML models on the three datasets.

### 4.4.1 Essay Classification

Given the large number of human-written essays collected in comparison to the LLMs, we decided to only use the first dataset. Despite that, we still had an unbalanced classification, as shown by dataset distribution in Figure 9.

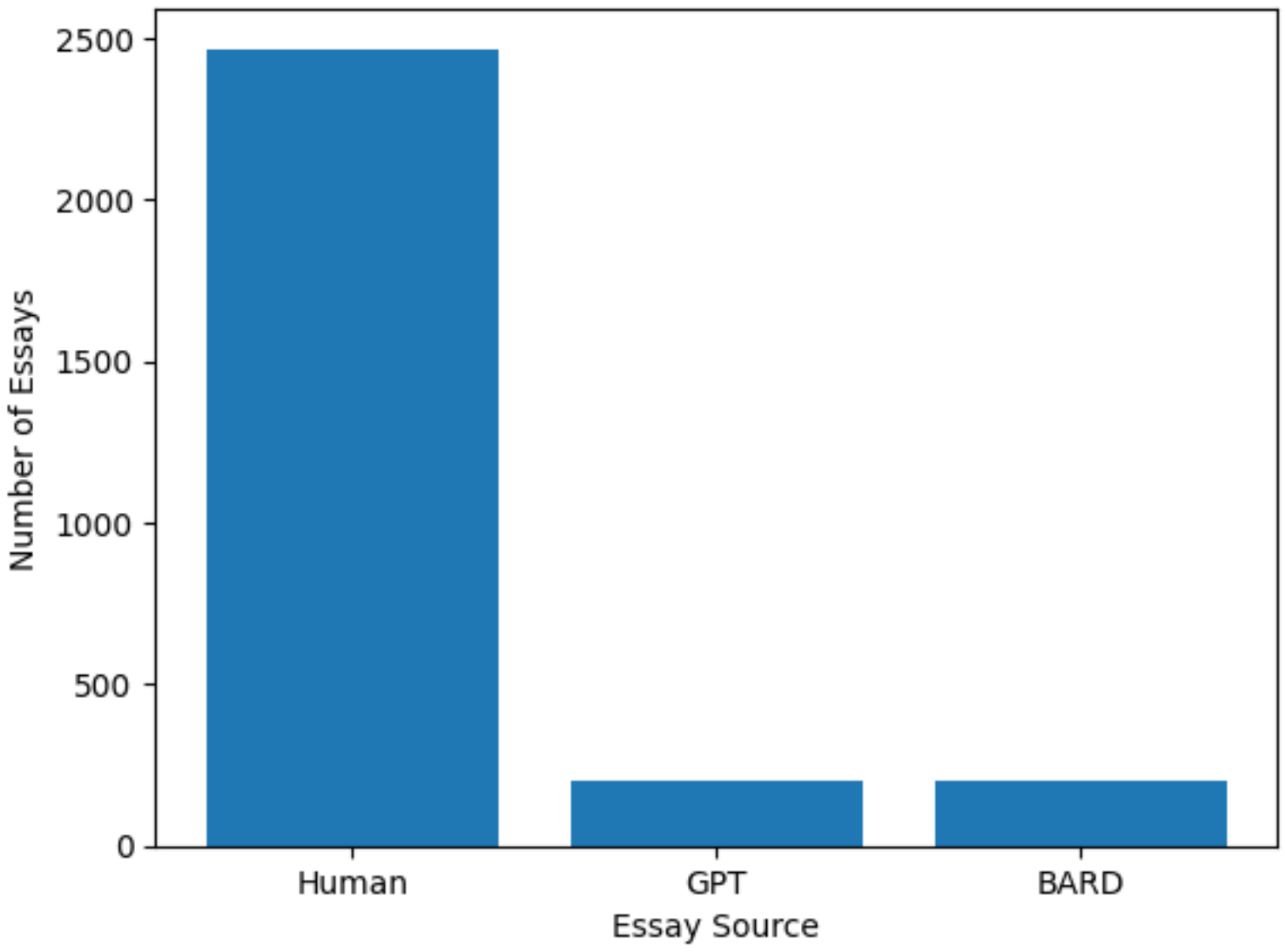

**Figure 9 Dataset Distribution for Essay Classification**

Table 1 displays the results for essay classification, averaged across the 5 folds. We note that all models perform relatively well when classifying human-written texts, likely because there are significantly more examples of this class in the dataset.

**Table 1 Essay Classification Results**

| Model | Accuracy | Precision | Recall | F1-Score |
|---|---|---|---|---|
| Random Forest | 0.9574 ± 0.0061 | 0.87 | 0.79 | 0.83 |
| SVM | 0.9452 ± 0.0061 | 0.74 | 0.74 | 0.74 |
| LR | 0.9340 ± 0.0064 | 0.72 | 0.68 | 0.70 |
| LSTM | 0.9414 ± 0.0110 | 0.73 | 0.73 | 0.73 |

The RF model delivered the best performance across all metrics. Its average accuracy was the highest, indicating that it correctly predicted the class for a higher percentage of essays than the other models. RF also had the highest F1 score, suggesting it was able to maintain a strong performance in both precision and recall. However, despite having the highest recall score, it still showed some limitations in accurately identifying all true positive instances. We present the confusion matrix from our best-performing model, RF, in Figure 10. Further results are presented in the Appendix.

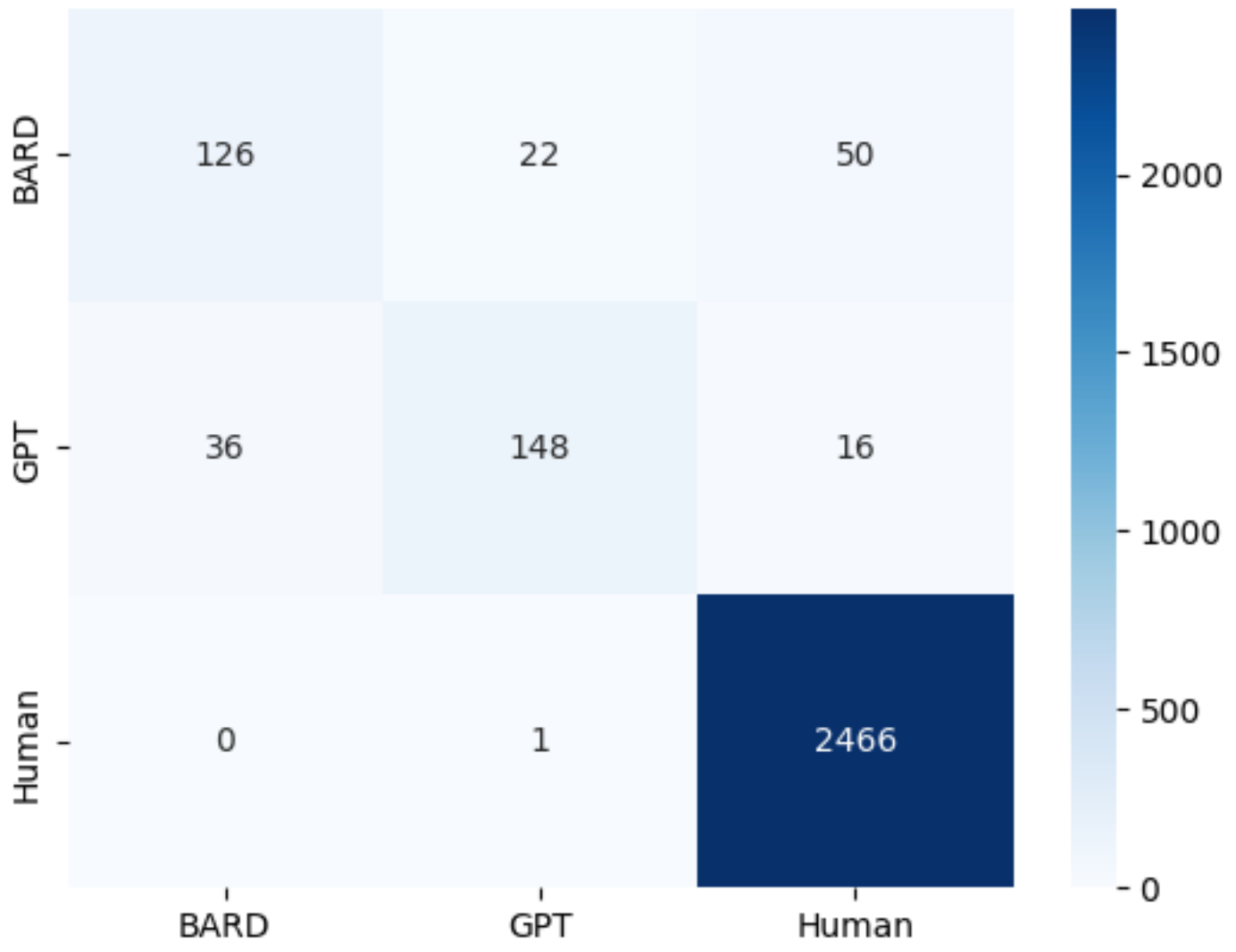

**Figure 10 Confusion Matrix for RF on Essay Classification**

The confusion matrix from the RF indicates it found it more challenging to correctly identify BARD-generated essays as compared to GPT-generated. The SVM had the next best performance but showed a slight decrease in all metrics compared to the RF model. The recall and F1-score were also lower, suggesting it had more difficulty correctly identifying all true positive instances and balancing precision and recall. The LR model demonstrated the weakest performance across all metrics. Finally, the standard deviation of accuracy for LSTM was higher, indicating its performance was more variable across different iterations.

In Figure 11, we display the dataset distribution for the story classification task. The distribution is more balanced compared to the previous dataset.

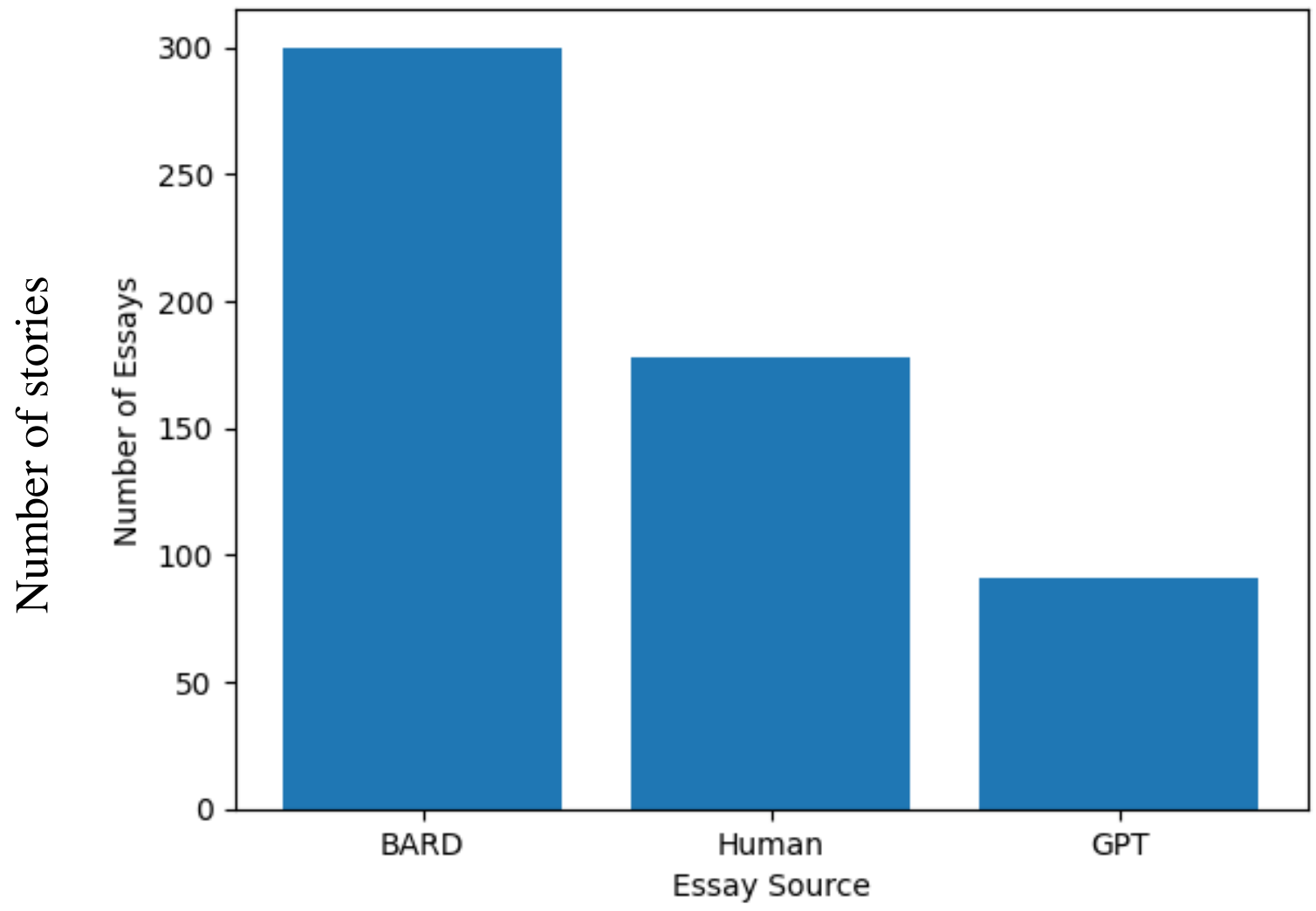

**Figure 11 Dataset Distribution for Story Classification**

Table 2 presents the result from story classification, averaged across the 5 folds. The results from the four models for story classification indicate variable performances. SVM provides the highest average accuracy of 88.73% with a standard deviation of 0.0198. The SVM also performs well in terms of average precision, recall, and F1 score, demonstrating a balance in the model's ability to minimize both false positives and false negatives. Both RF and LR models display a lower but relatively similar level of accuracy, with 84.50% and 85.39%, respectively. However, their average precision values are higher than SVM and LSTM, which indicates a lower rate of false positives. Finally, the LSTM model has the lowest accuracy, precision, recall, and F1 scores among all four models. The LSTM model may need more training data or further tuning of the hyperparameters to improve performance. Figure 12 illustrates the confusion matrix for the best performing SVM model. We present additional results in the Appendix.

**Table 2 Story Classification Results**

| Model | Accuracy | Precision | Recall | F1-Score |
|---|---|---|---|---|
| RF | 0.8450 (± 0.0135) | 0.87 | 0.70 | 0.69 |
| SVM | 0.8873 (± 0.0198) | 0.85 | 0.80 | 0.81 |
| LR | 0.8539 (± 0.0162) | 0.87 | 0.71 | 0.69 |
| LSTM | 0.8240 (± 0.0120) | 0.71 | 0.65 | 0.59 |

For detecting BARD (Class 0) stories, all models demonstrated high precision, but recall and F1 scores varied; SVM exhibited the best performance with a 0.99 F1 score. Detecting GPT (Class 1) stories proved challenging, with SVM and LR displaying high precision but low recall, suggesting a higher rate of missed instances; LSTM exhibited a high rate of false positives. Conversely, all models excelled in detecting human-written stories (Class 2), demonstrating high precision and recall; SVM had the highest F1 score (0.84), showing a robust balance between precision and recall. Overall, the models effectively identified BARD and Human classes but struggled with GPT.

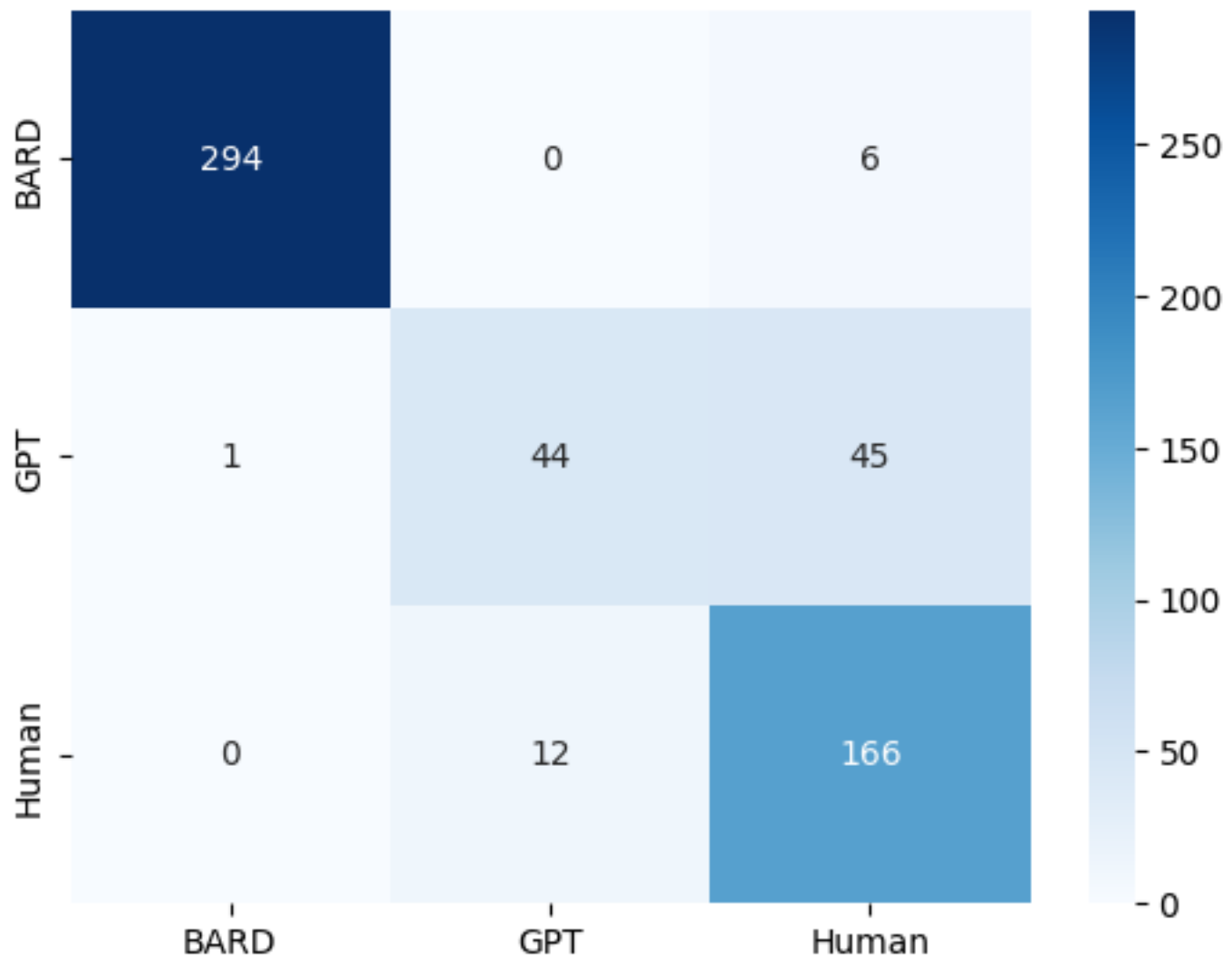

**Figure 12 Confusion Matrix for SVM on Story Classification**

In Figure 13, we present the dataset distribution for poetry classification. Despite randomly sampling 50% of the poems from the dataset, we still had an imbalanced classification problem.

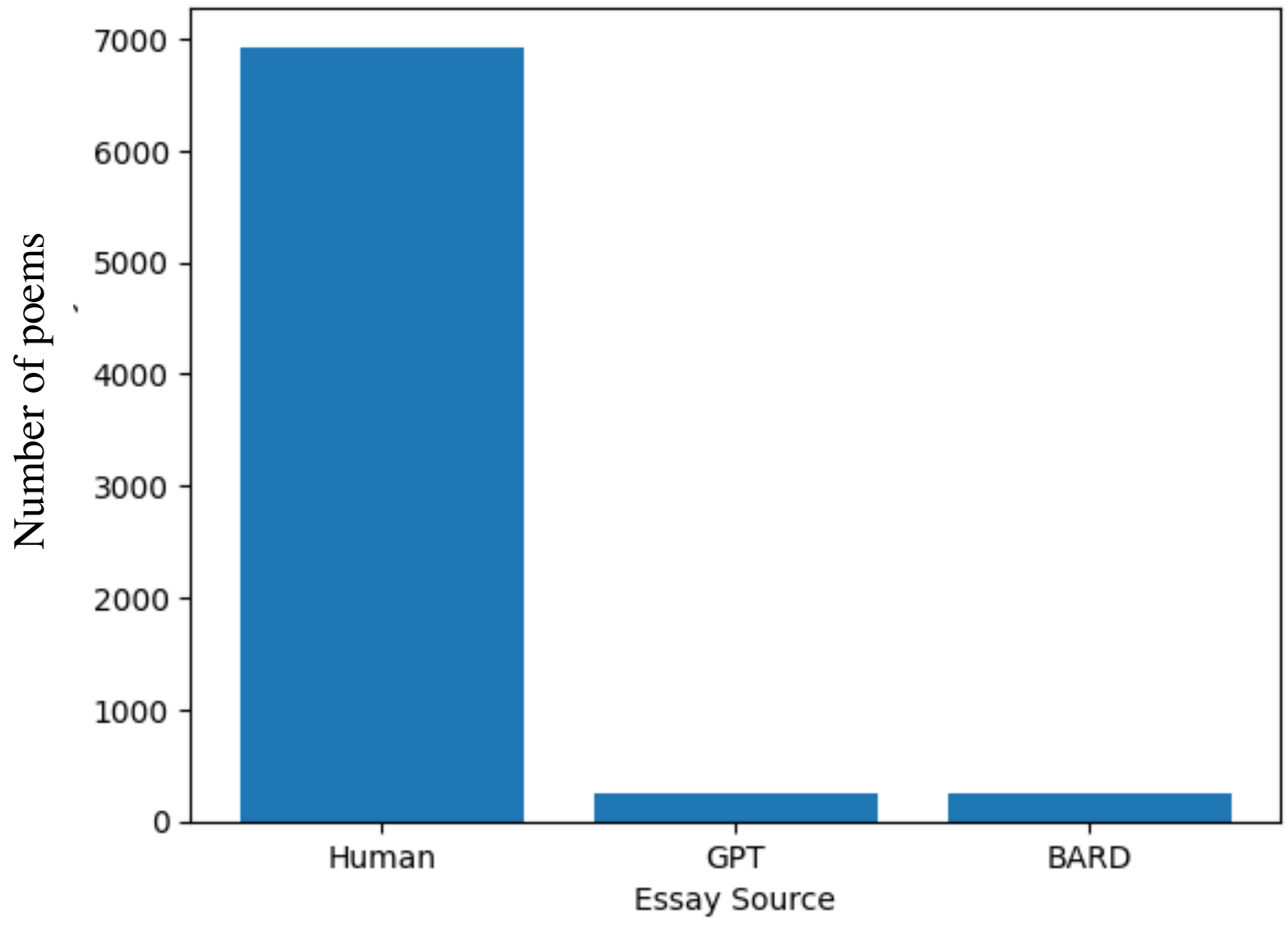

**Figure 13 Dataset Distribution for Poetry Classification**

Table 3 displays the result from poetry classification, averaged across 5 folds. SVM emerged as the most effective model with an accuracy of 0.9842 and an F1-score of 0.89. The RF model had the lowest performance metrics across all categories, with only 0.42 F1-score. The class imbalance, with the majority of poems being human-written, likely significantly influenced these results. In Figure 14, we present the confusion matrix for the best-performing SVM model on poem classification. Additional results are presented in the Appendix.

**Table 3 Poetry Classification Results**

| Model | Accuracy | Precision | Recall | F1-Score |
|---|---|---|---|---|
| RF | 0.9376 ± 0.0014 | 0.95 | 0.39 | 0.42 |
| SVM | 0.9842 ± 0.0028 | 0.94 | 0.85 | 0.89 |
| LR | 0.9661 ± 0.0036 | 0.95 | 0.67 | 0.76 |
| LSTM | 0.9629 ± 0.0047 | 0.68 | 0.65 | 0.65 |

Class 2, representing human-written poetry, showed the highest precision, recall, and F1-score across all models, emphasizing the models' ability to identify human-written poetry. However, the models showed more variation when classifying LLM-generated poetry, with BARD (class 0) and GPT (class 1). For BARD, the SVM model performed the best with a precision of 0.91 and recall of 0.77, indicating reasonable success in correctly detecting BARD-generated poetry. However, LSTM struggled the most with this class, exhibiting the lowest precision and recall values. As for GPT (class 1), SVM performed the best, having equal precision and recall values as for BARD. Despite LSTM's struggle with classifying BARD-generated poetry, it was better at classifying GPT-generated poetry.

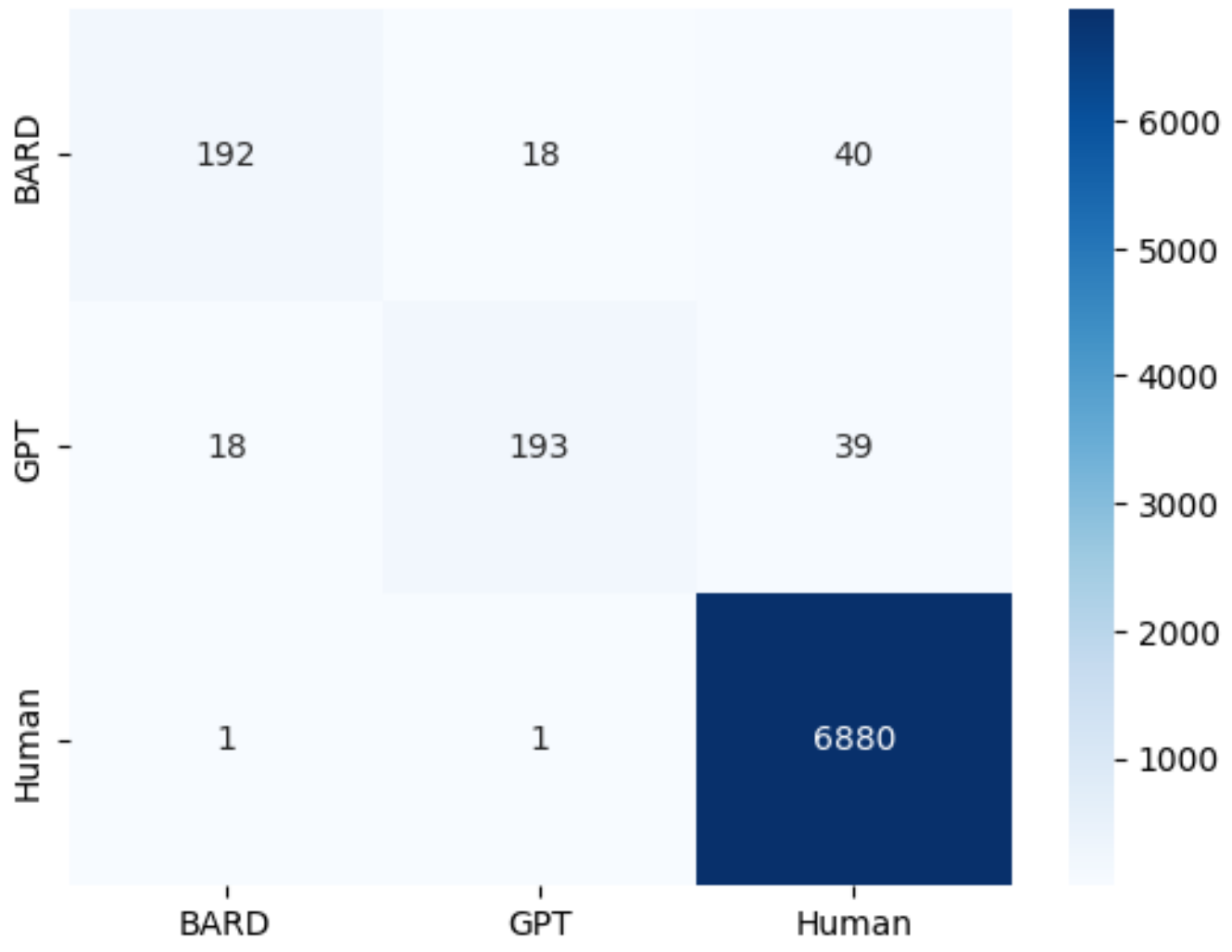

**Figure 14 Confusion Matrix for SVM on Poem Classification**

## *4.5 Results on Binary classification*

Table 4 displays the results for essay classification, averaged across 5 folds. In the Human vs GPT classification task, SVM exhibited the best performance, with an accuracy of 99.85% with a minimal standard deviation. It also scored the highest in precision, recall, and F1-score, denoting its effective detection of both classes. For the Human vs BARD classification, the SVM model again performed the best, with an accuracy of 99.66%. The lower recall of LR indicates that while it accurately identified the positive class (human-authored text), it struggled with BARD-generated text, leading to higher false negative instances. Figures 15 and 16 illustrate the confusion matrix for the best-performing SVM model. Overall, all the models perform exceptionally well with over 90% F1-score in both tasks.

**Table 4 Essay Binary Classification Results**

| Model | Accuracy | Precision | Recall | F1-Score |
|---|---|---|---|---|
| | | **Human vs GPT** | | |
| RF | 0.9918 ± 0.0025 | 0.99 | 0.95 | 0.97 |
| SVM | 0.9985 ± 0.0014 | 1.00 | 0.99 | 0.99 |
| LR | 0.9854 ± 0.0051 | 0.99 | 0.90 | 0.94 |
| LSTM | 0.9940 ± 0.0052 | 0.99 | 0.97 | 0.98 |
| | | **Human vs BARD** | | |
| RF | 0.9861 ± 0.0035 | 0.99 | 0.91 | 0.94 |
| SVM | 0.9966 ± 0.0008 | 1.00 | 0.98 | 0.99 |
| LR | 0.9775 ± 0.0039 | 0.99 | 0.85 | 0.90 |
| LSTM | 0.9914 ± 0.0025 | 0.99 | 0.95 | 0.97 |

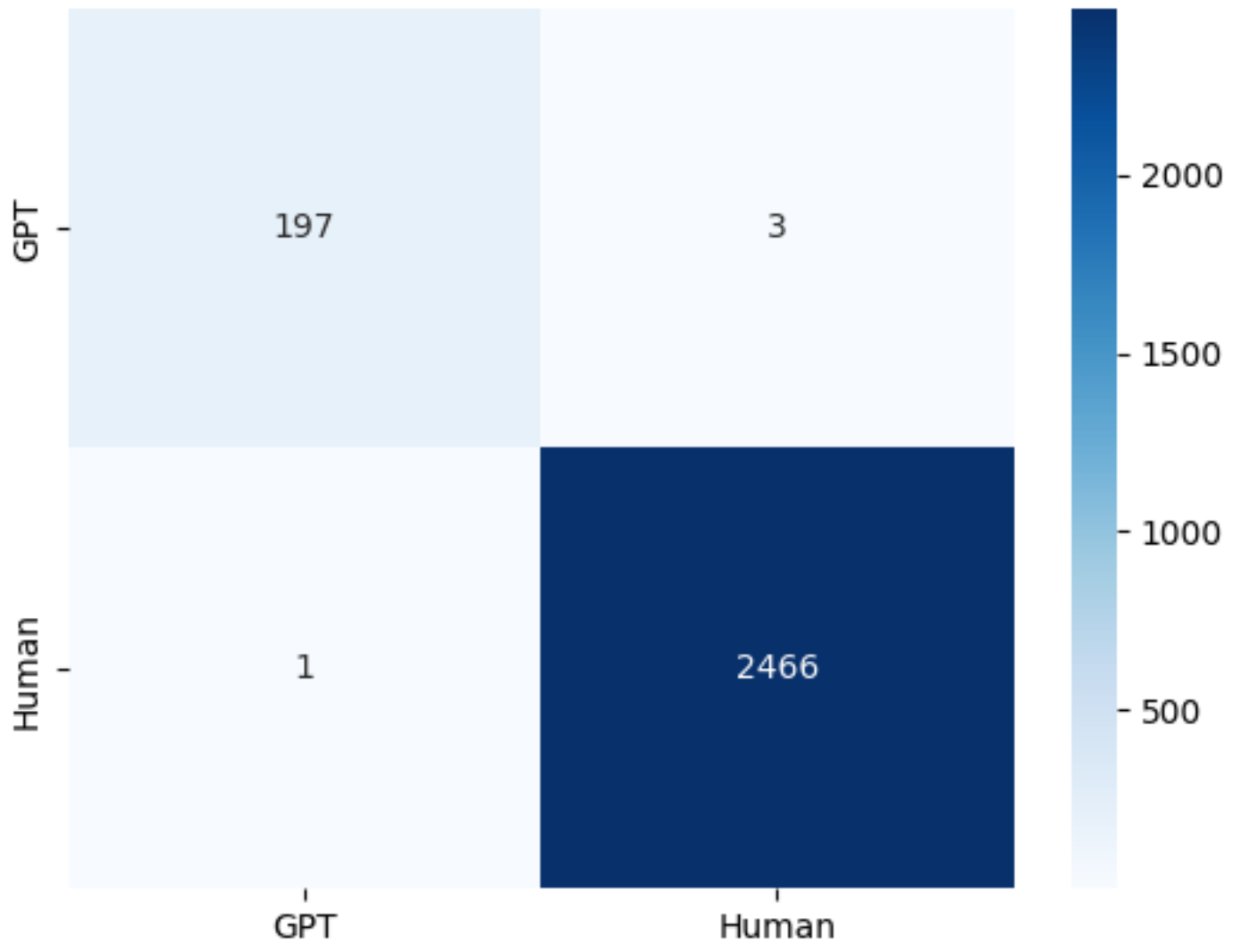

**Figure 15 Confusion Matrix for SVM on Classifying human and GPT essays**

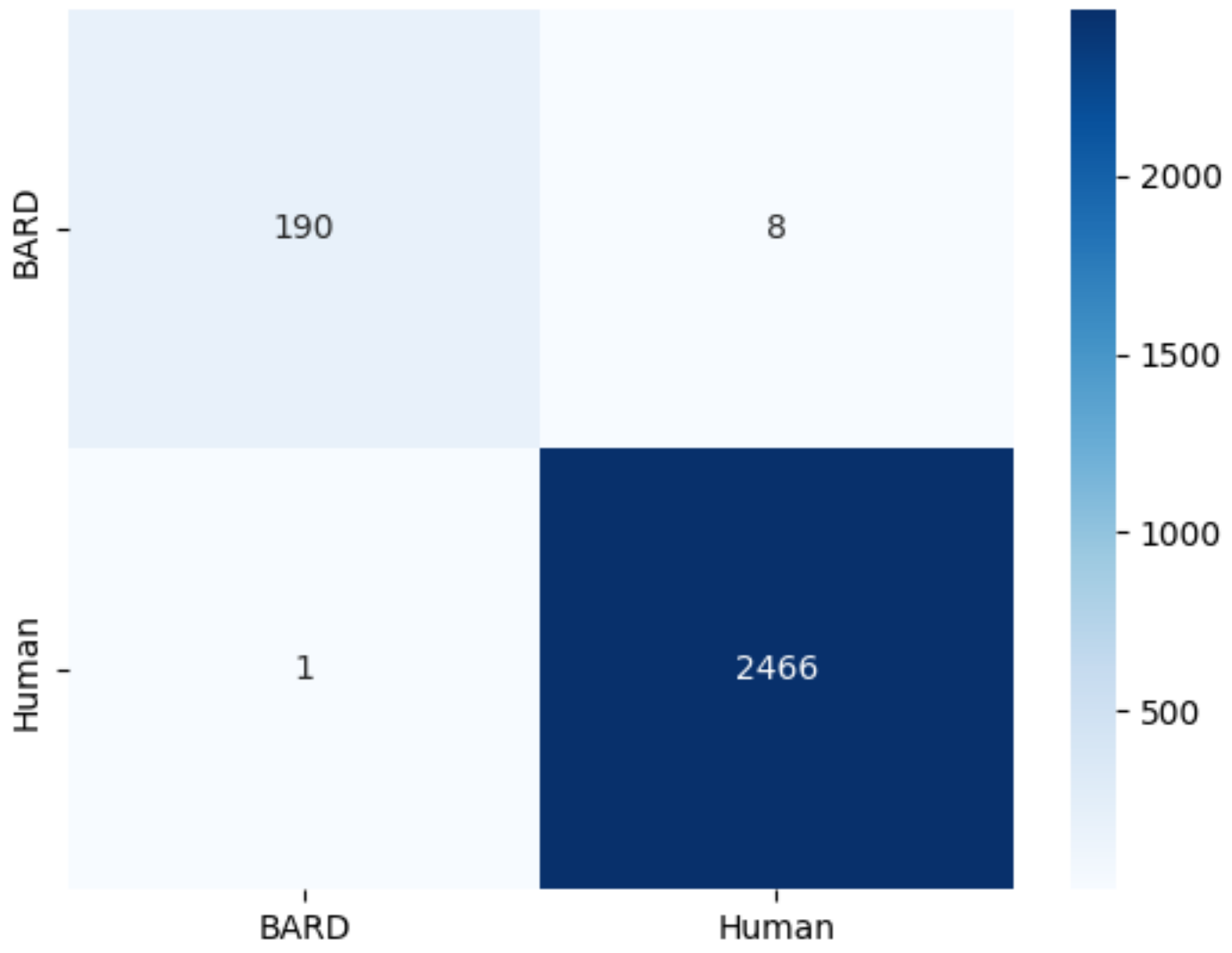

**Figure 16 Confusion Matrix for SVM on Classifying human and BARD essays**

Table 5 displays the results for story classification, averaged across 5 folds. In distinguishing human-generated stories from those generated by GPT, SVM outperformed the other models with the highest accuracy and balanced precision (0.80), recall (0.73), and F1-score (0.75). The LSTM model struggled, with the lowest scores, likely due to a failure to predict GPT-generated stories. In contrast, when differentiating human-generated stories from BARD-generated, all models performed exceptionally well. Thus, BARD stories were easier to detect for the ML models than GPT stories. All the models performed poorly in the former task, with 0.75 being the highest F1-score they all performed exceptionally well in the latter, with more than 0.93 F1-score across all models. Figures 17 and 18 illustrate the confusion matrix for the best-performing SVM model.

**Table 5 Story Classification Results (Binary)**

| Model | Accuracy | Precision | Recall | F1-Score |
|---|---|---|---|---|
| **Human vs GPT** | | | | |
| RF | 0.7536 ± 0.0308 | 0.84 | 0.63 | 0.63 |
| SVM | 0.7983 ± 0.0332 | 0.80 | 0.73 | 0.75 |
| LR | 0.6828 ± 0.0121 | 0.84 | 0.53 | 0.46 |
| LSTM | 0.6642 ± 0.0031 | 0.33 | 0.50 | 0.40 |
| **Human vs BARD** | | | | |
| RF | 0.9812 ± 0.0138 | 0.98 | 0.98 | 0.98 |
| SVM | 0.9875 ± 0.0078 | 0.98 | 0.99 | 0.99 |
| LR | 0.9875 ± 0.0102 | 0.99 | 0.99 | 0.99 |
| LSTM | 0.9477 ± 0.0355 | 0.95 | 0.94 | 0.94 |

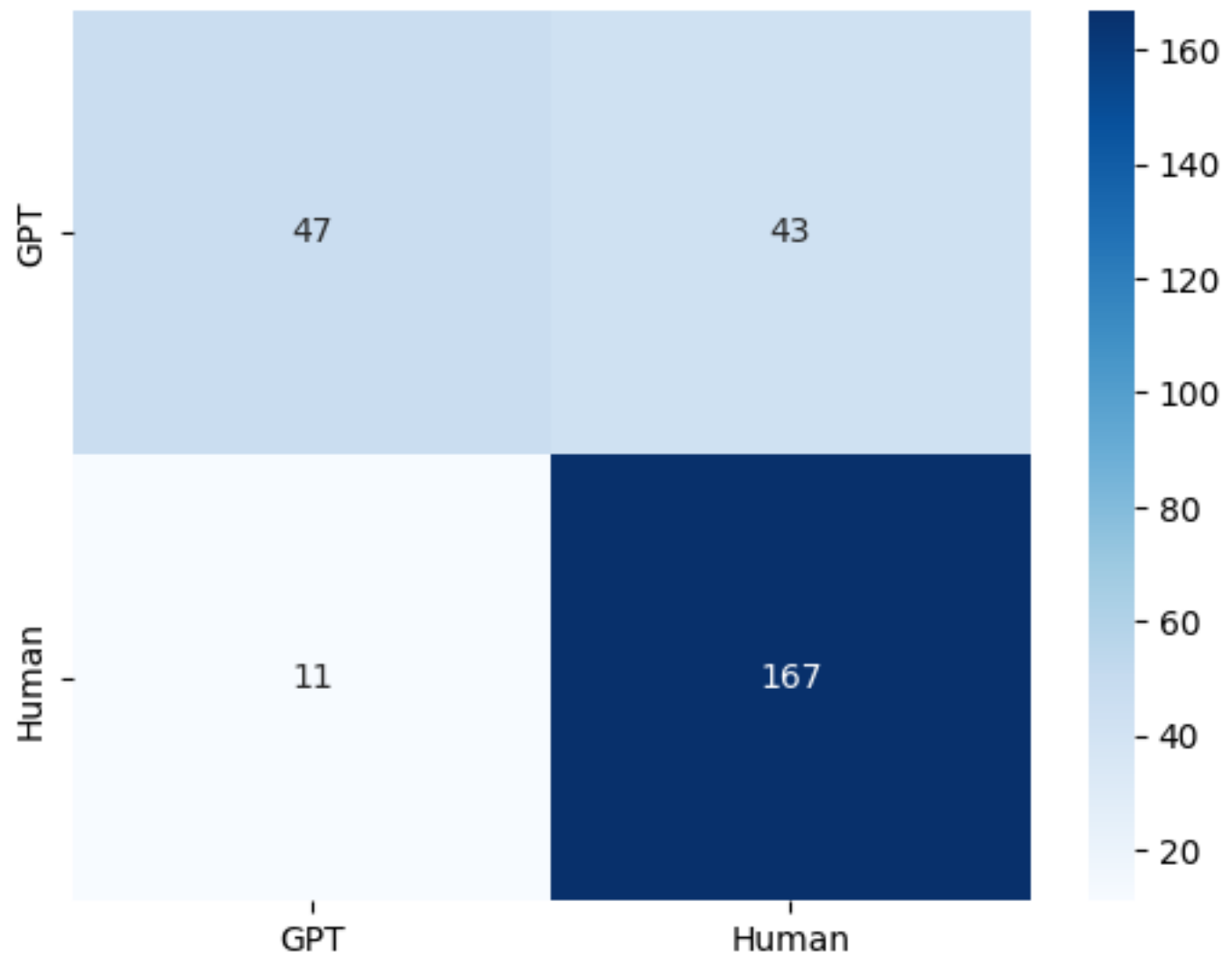

**Figure 17 Confusion Matrix for SVM on Classifying human and GPT stories**

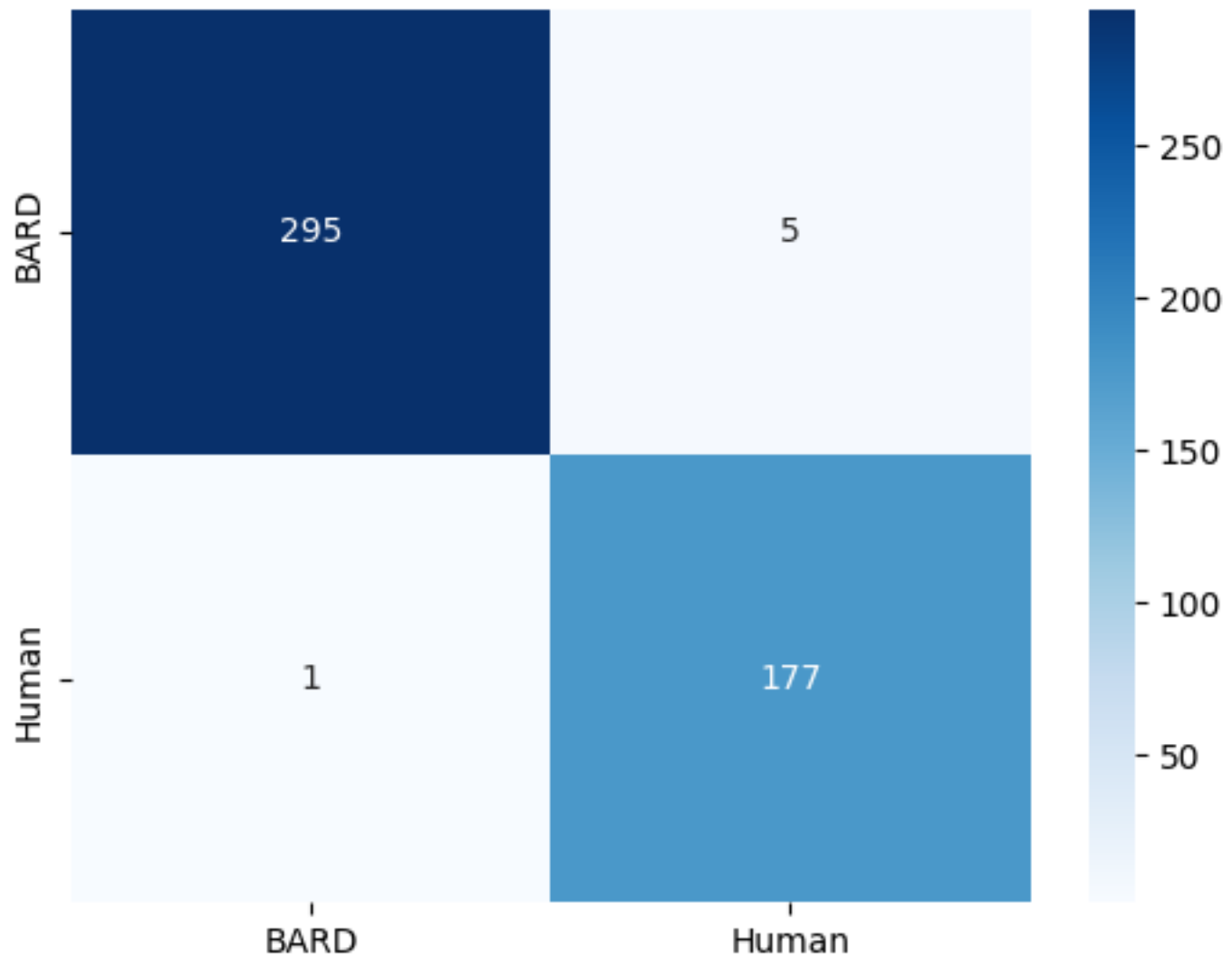

**Figure 18 Confusion Matrix for SVM on Classifying human and BARD stories**

Table 6 displays the results for poetry classification, averaged across 5 folds. SVM showed the best overall performance, closely followed by LSTM. SVM generally outperforms the other models across both tasks. LSTM also shows strong performance, particularly in its balance across precision, recall, and F1-score. RF and LR, on the other hand, struggle particularly with recall, indicating difficulty in correctly identifying all instances of the positive class. Figures 19 and 20 illustrate the confusion matrix for the best-performing SVM model.

**Table 6 Poetry Classification Results (Binary)**

| Model | Accuracy | Precision | Recall | F1-Score |
|---|---|---|---|---|
| | | **Human vs GPT** | | |
| RF | 0.9675 ± 0.0009 | 0.98 | 0.53 | 0.55 |
| SVM | 0.9941 ± 0.0014 | 0.99 | 0.92 | 0.95 |
| LR | 0.9752 ± 0.0007 | 0.99 | 0.65 | 0.72 |
| LSTM | 0.9912 ± 0.0032 | 0.96 | 0.90 | 0.93 |
| | | **Human vs BARD** | | |
| RF | 0.9672 ± 0.0014 | 0.98 | 0.53 | 0.54 |
| SVM | 0.9931 ± 0.0026 | 0.99 | 0.90 | 0.94 |
| LR | 0.9764 ± 0.0019 | 0.99 | 0.66 | 0.74 |
| LSTM | 0.9885 ± 0.0091 | 0.92 | 0.90 | 0.91 |

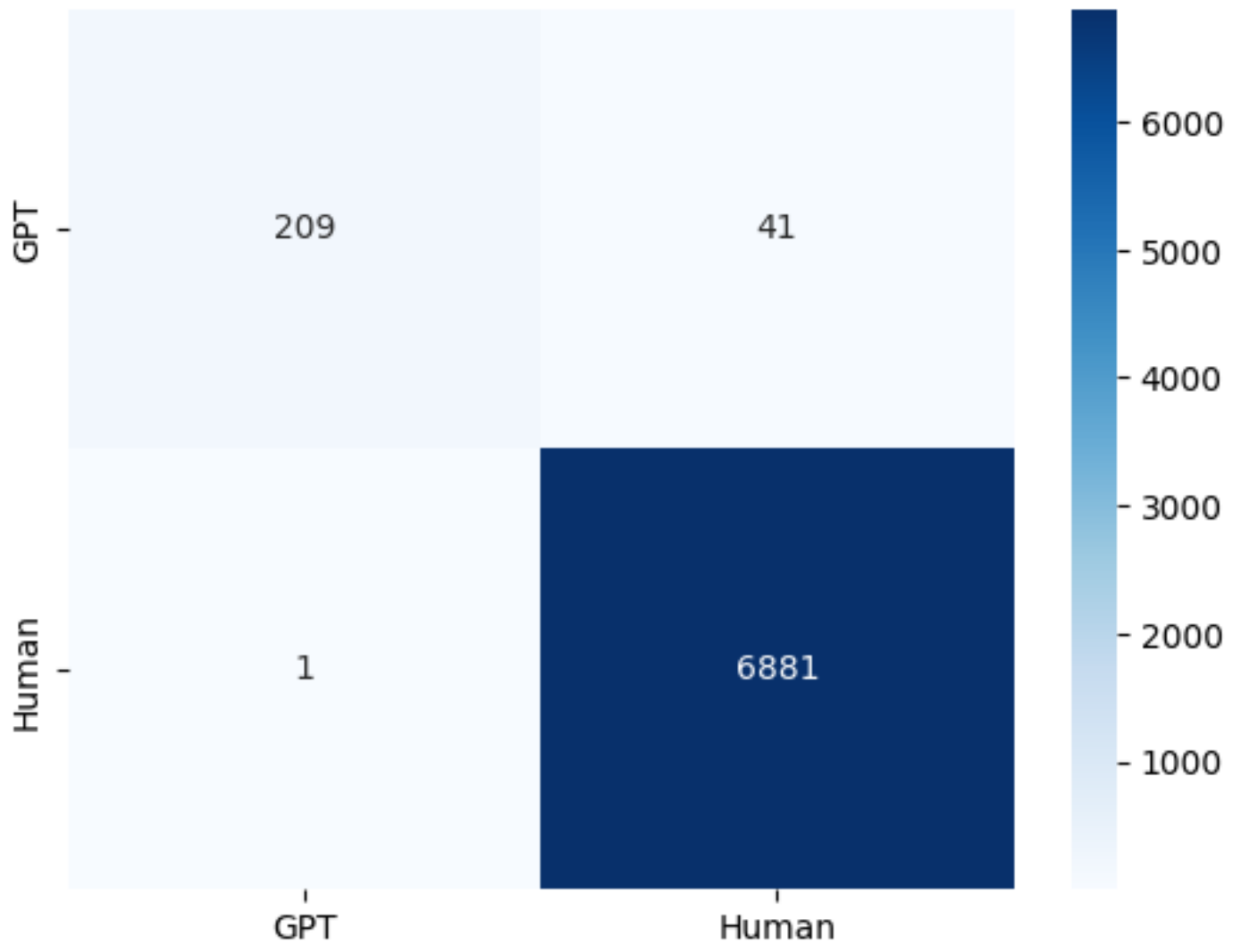

**Figure 19 Confusion Matrix for SVM on Classifying human and GPT poems**

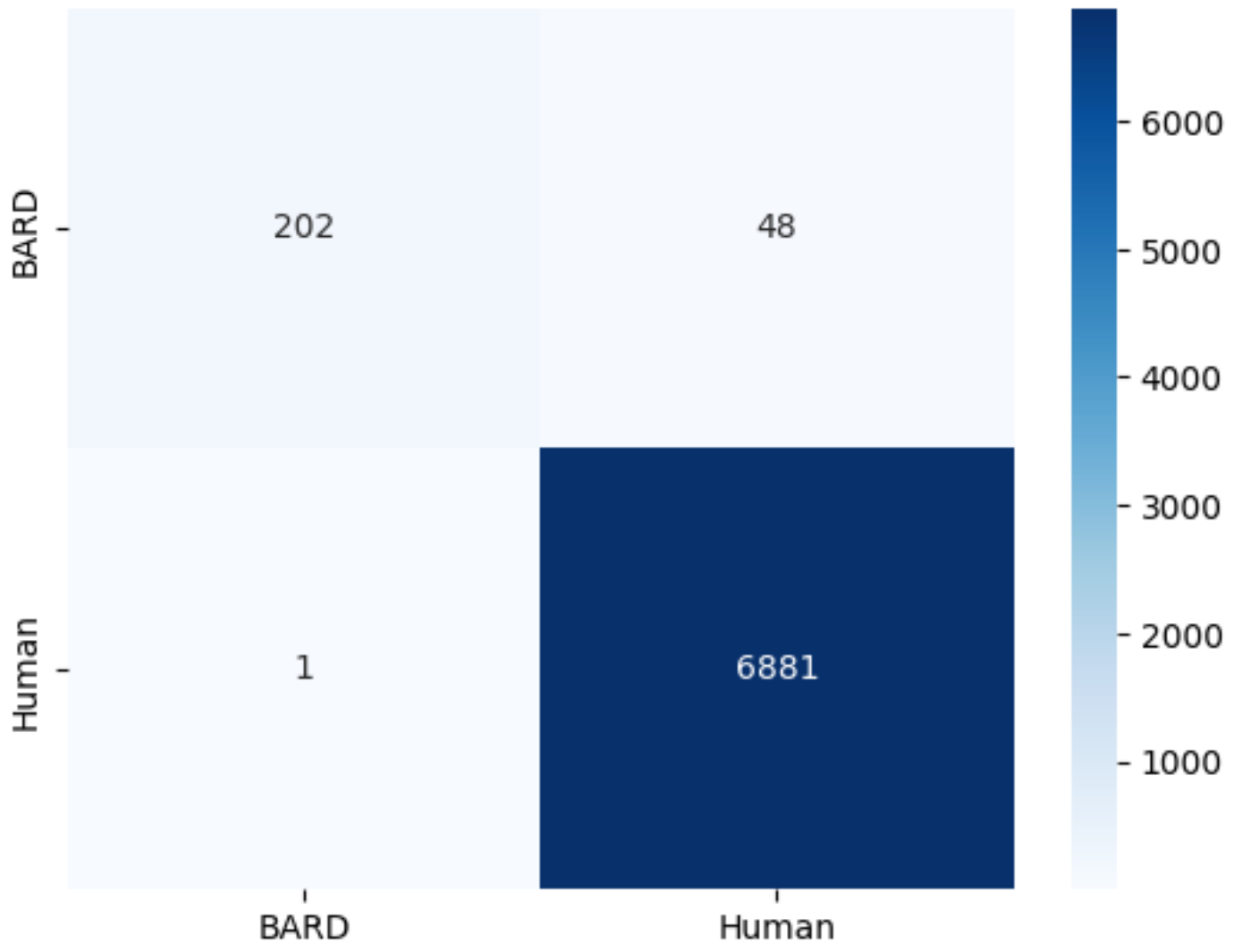

**Figure 20 Confusion Matrix for SVM on Classifying human and BARD poems**

## 5. Discussion

In this section, we discuss potential impacts of our work. We also highlight some of the limitations and hint at future research directions.

### *5.1 Implications*

This research has several significant implications, particularly in differentiating between human-generated and AI-generated texts. Firstly, the relatively high performance achieved across most experiments, especially SVM, in distinguishing between human and GPT-generated or BARD-generated text indicates that with the right features and model configuration, it's feasible to detect artificially generated text. The only exception is the poor performance of the models in detecting GPT-generated stories from human-written ones. The results could extend to various applications, such as detecting deep fake text in news or social media or validating the authenticity of content. The results also suggest that ML can successfully differentiate between the nuances in human writing and algorithmic output, even with limited data size. However, the differences between the performance of the classifiers when identifying GPT or BARD versus human text, suggest that different AI writing models might have unique "fingerprints" or characteristics that these classifiers can detect. This could lead to the development of more specialized detection tools for different types of LLMs. Finally, the study also contributes to raising ethical implications in the context of LLMs in academic integrity. As AI text generation improves, the potential for misuse increases, necessitating the importance of effective detection tools. It also emphasizes the need for a discussion about the responsibility of AI developers, researchers, and users in ensuring these technologies are used ethically and safely.

### *5.2 Future Work*

One of the future research directions could be exploring the influence of feature selection on the classifiers' performance. It could be beneficial to delve deeper into the differentiating characteristics that the classifiers are detecting and determine whether these can be refined or augmented for superior performance. The study was limited to identifying text generated from two LLMs, GPT and BARD. Therefore, future work could extend the comparison to include other LLM models like Falcon [38] and CTRL [39]. This could provide a better understanding of their unique features and the ease with which they can be differentiated from human text. Moreover, extending this analysis to other languages besides English would be valuable. It would be interesting to see if the same techniques and models are effective in different linguistic contexts, or if language-specific models need to be developed. Although we collected Python code from GPT and BARD, there is a need to integrate human-written python3 code into the dataset and evaluate the effectiveness of ML models in categorizing software codes. Finally, alongside improving detection tools, it's vital to advance discussions on ethical guidelines and policies to manage the use of increasingly sophisticated AI text generators.

## 6. Conclusion

This research marks a step forward in the field of LLMs, specifically in distinguishing human-written text from AI-generated content. A key contribution of this work is the introduction of a novel dataset for evaluating LLM-generated text from GPT and BARD across multiple genres, including stories and poetry. Our results indicate that while LLMs have made remarkable progress, it remains possible to discern human-written texts from those generated by LLMs using ML and DL models with a small training set. We observed high performance when differentiating BARD-generated texts from human-written texts. However, the task becomes more challenging with GPT-generated text, especially in story writing. By creating and leveraging this unique dataset, we have not only deepened our understanding of the nuanced differences between LLM and human writing but also provided a valuable resource for future research and development in this emerging field.

## Author statements

*Ethical Approval*

Ethical approval was not required because no personal data was used. Any analysis presented were aggregated.

*Competing Interests*

None declared.

# Appendix

*Multiclass Classification*

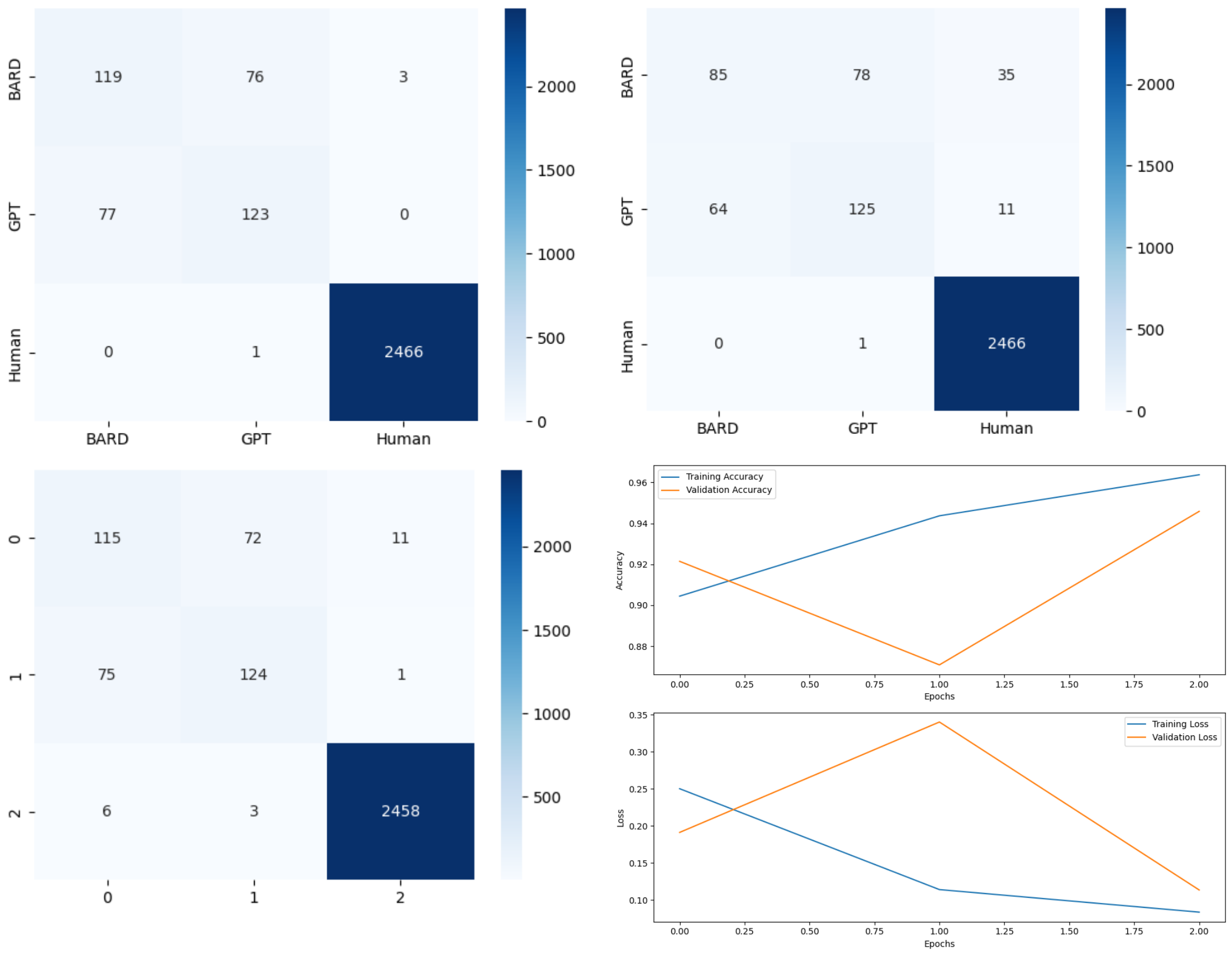

**Figure 21 Confusion Matrix for SVM (top left), LR (top right), LSTM (bottom left), and Training Curve for LSTM (bottom right) on multiclass essay classification**

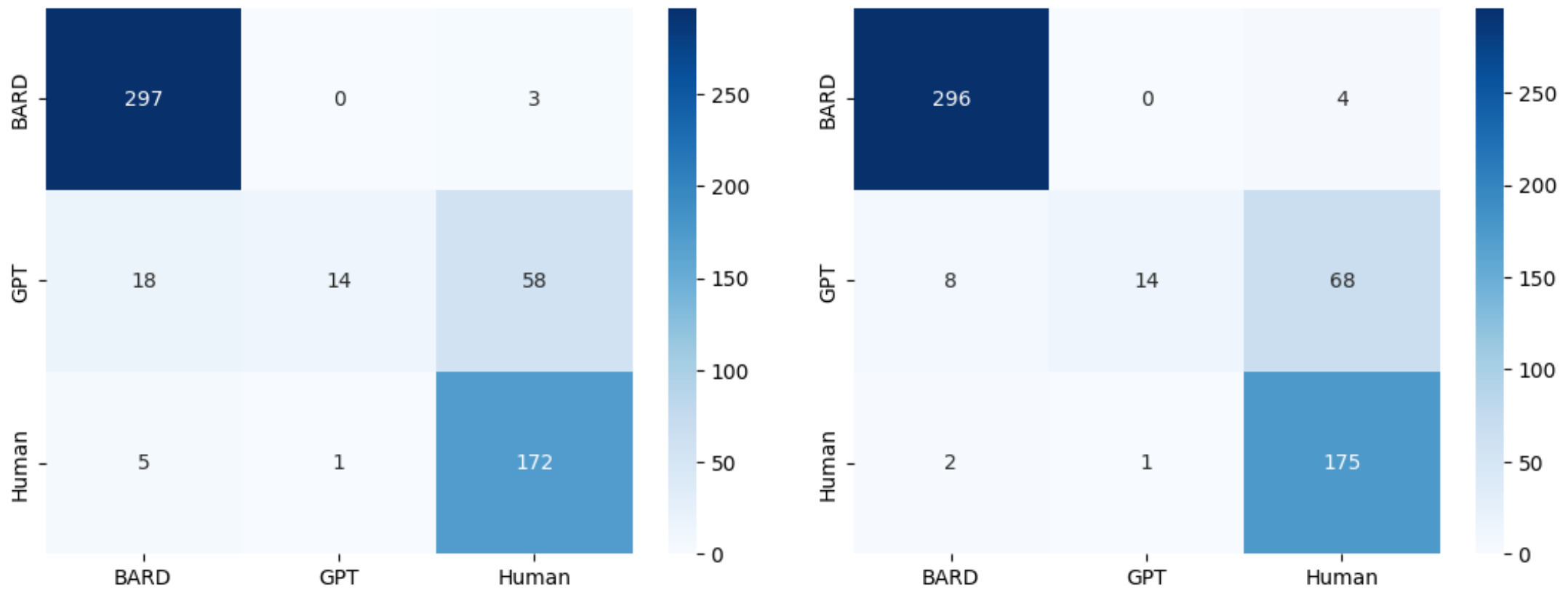

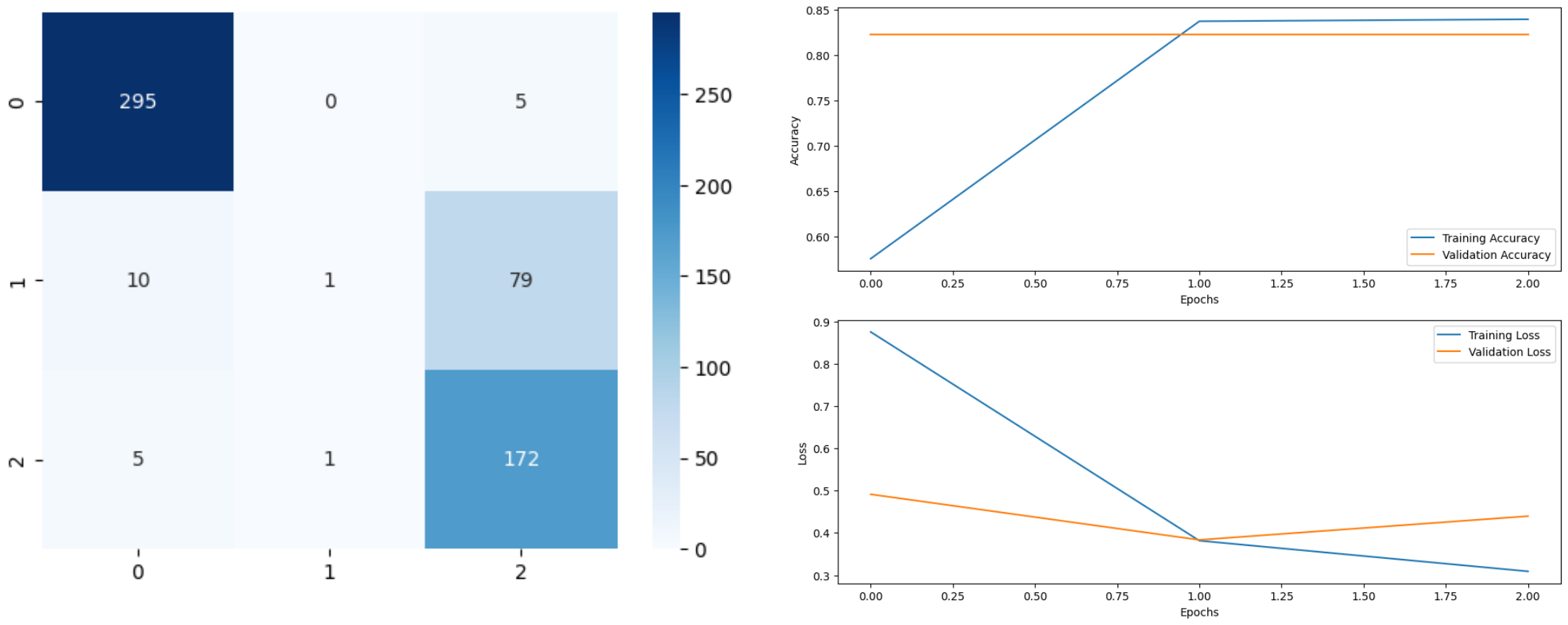

**Figure 22 Confusion Matrix for RF (top left), LR (top right), LSTM (bottom left), and Training Curve for LSTM (bottom right) on multiclass story classification**

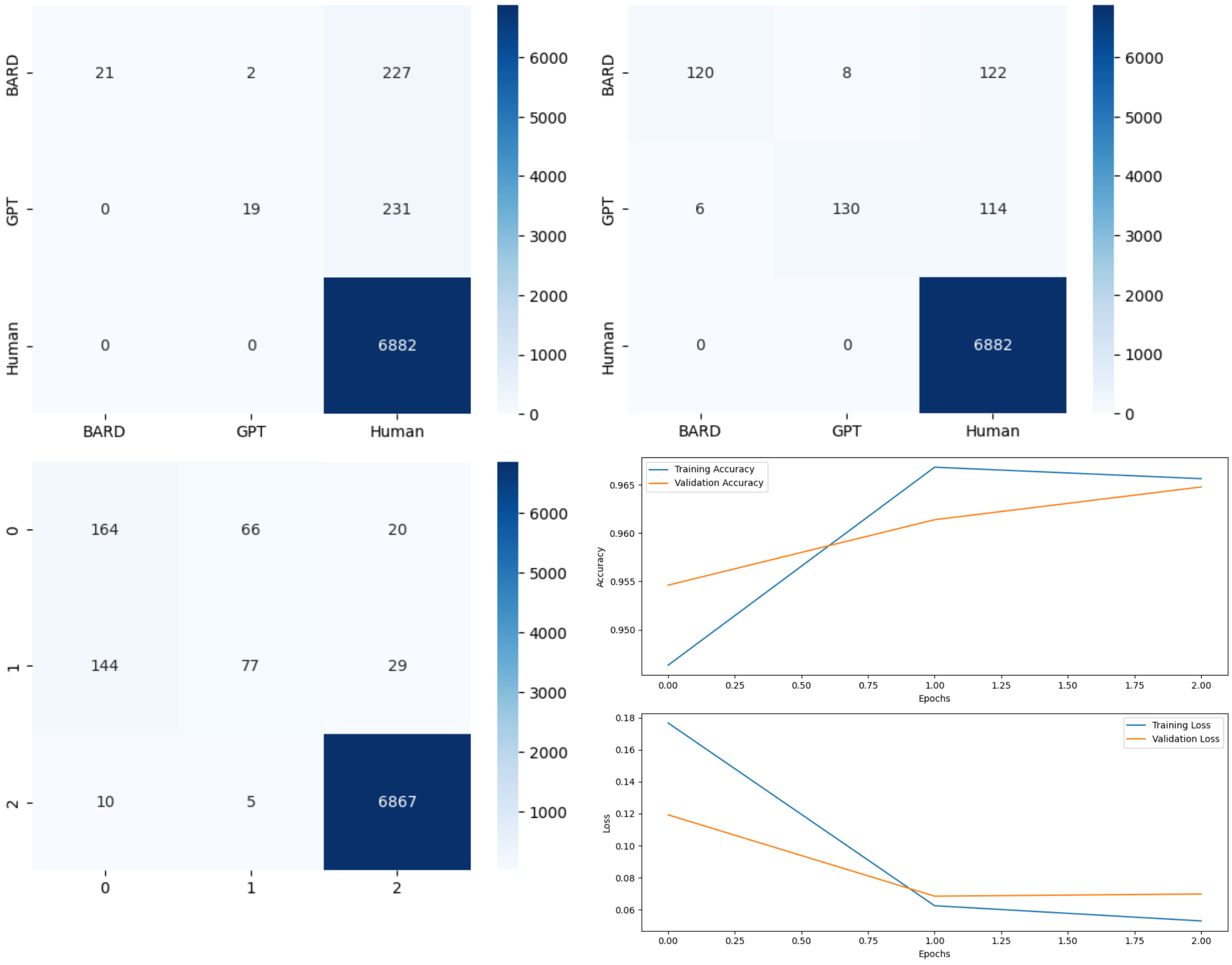

**Figure 23 Confusion Matrix for RF (top left), LR (top right), LSTM (bottom left), and Training Curve for LSTM (bottom right) on multiclass poetry classification**

*Binary Classification*

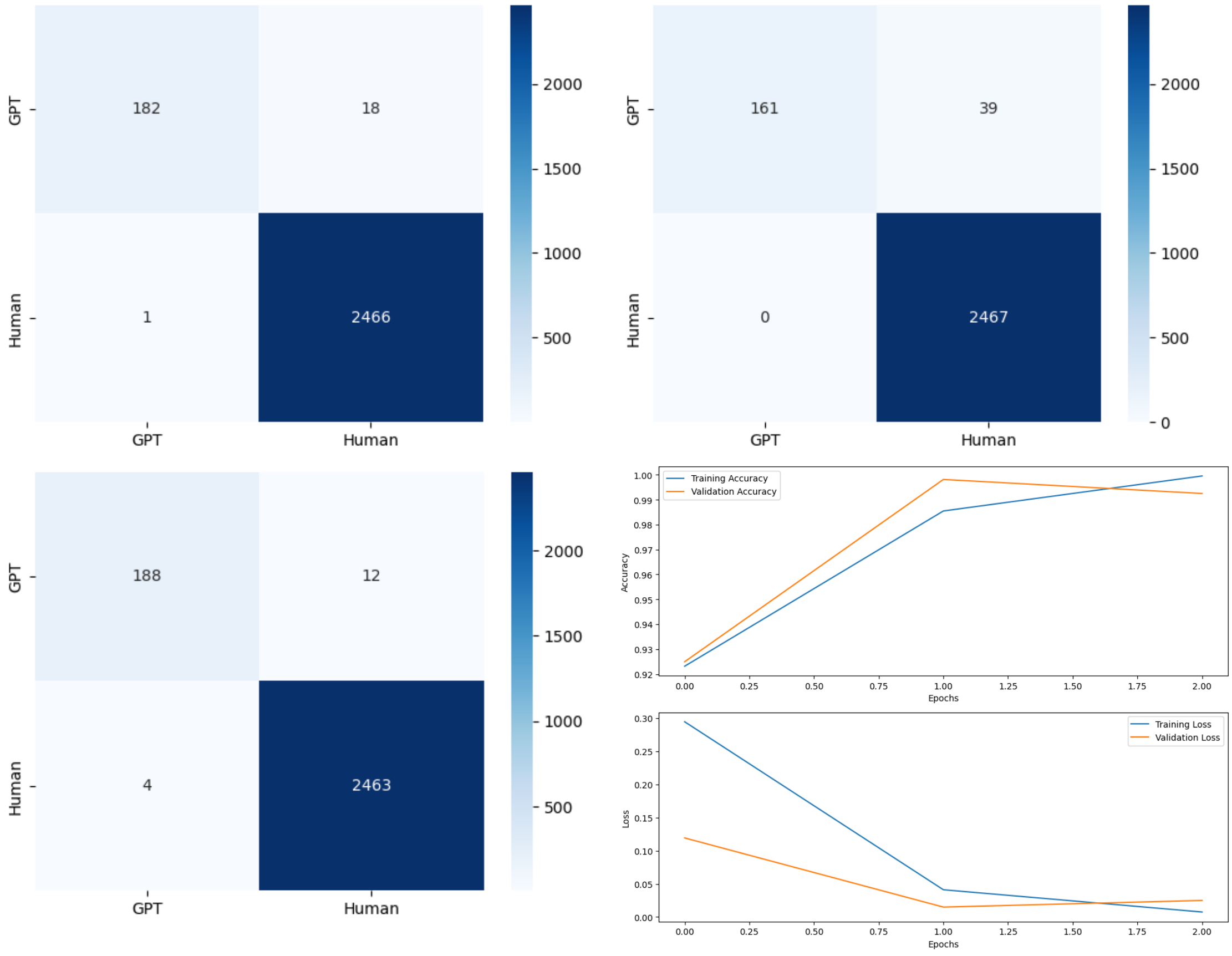

**Figure 24 Confusion Matrix for RF (top left), LR (top right), LSTM (bottom left), and Training Curve for LSTM (bottom right) on binary essay classification (Human vs GPT)**

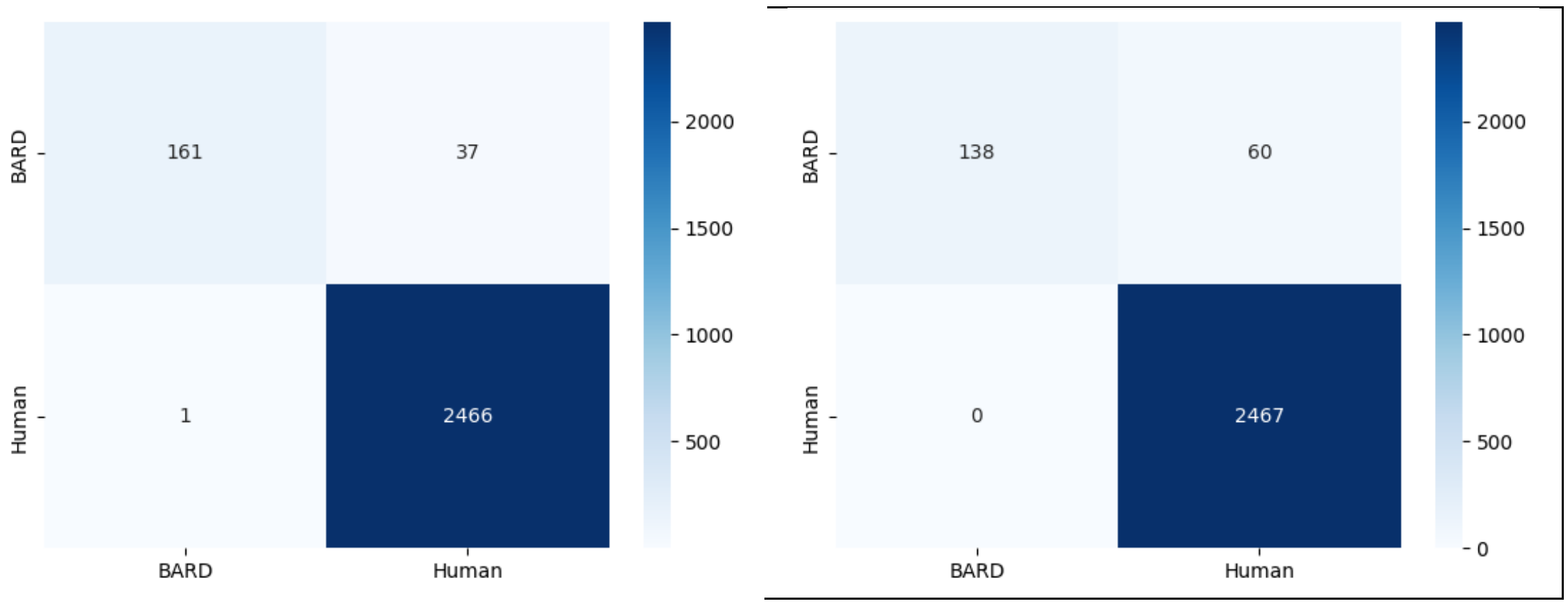

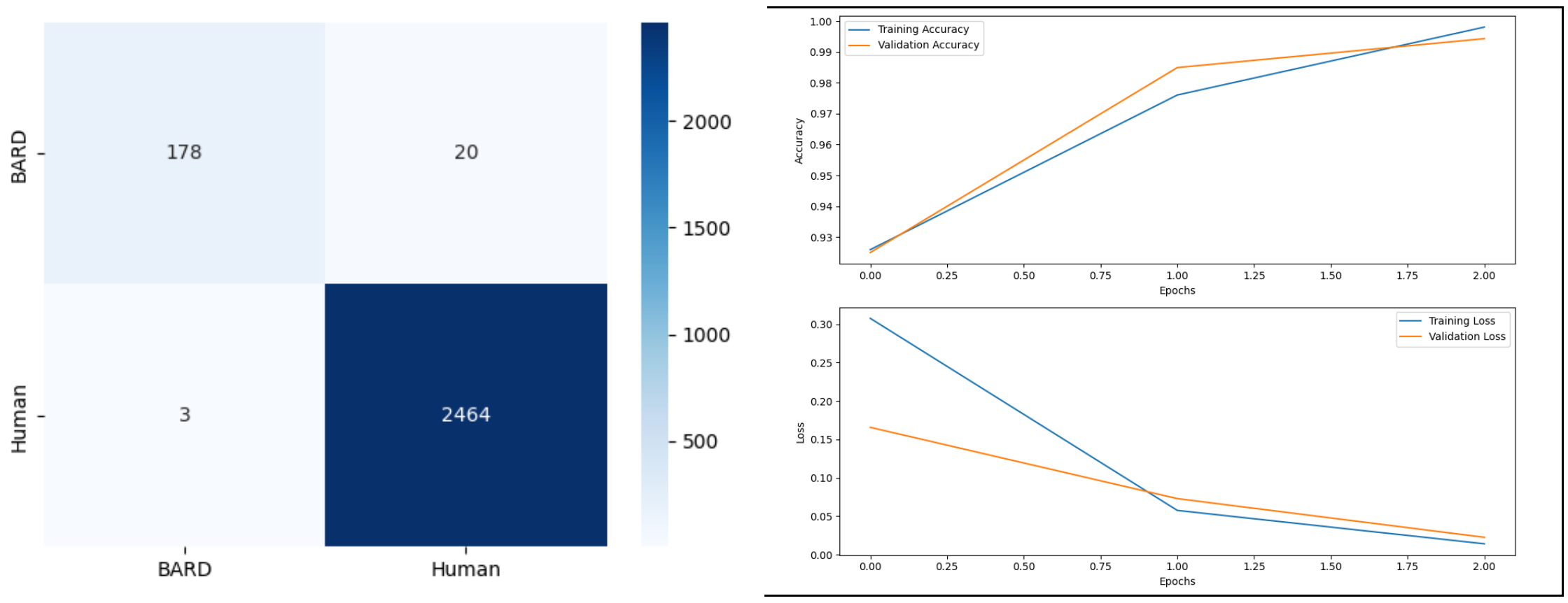

**Figure 25 Confusion Matrix for RF (top left), LR (top right), LSTM (bottom left), and Training Curve for LSTM (bottom right) on binary essay classification (Human vs BARD)**

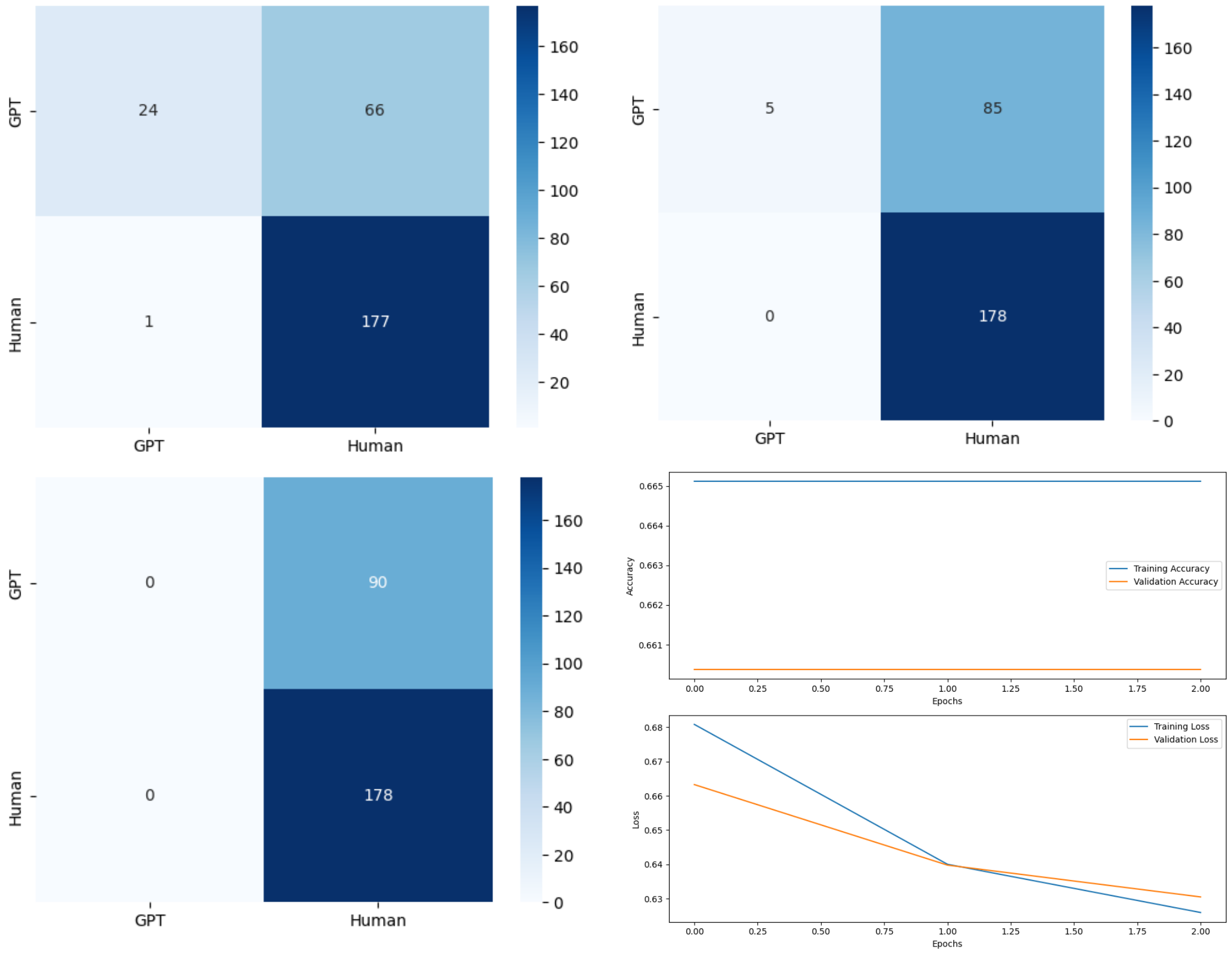

**Figure 26 Confusion Matrix for RF (top left), LR (top right), LSTM (bottom left), and Training Curve for LSTM (bottom right) on binary story classification (Human vs GPT)**

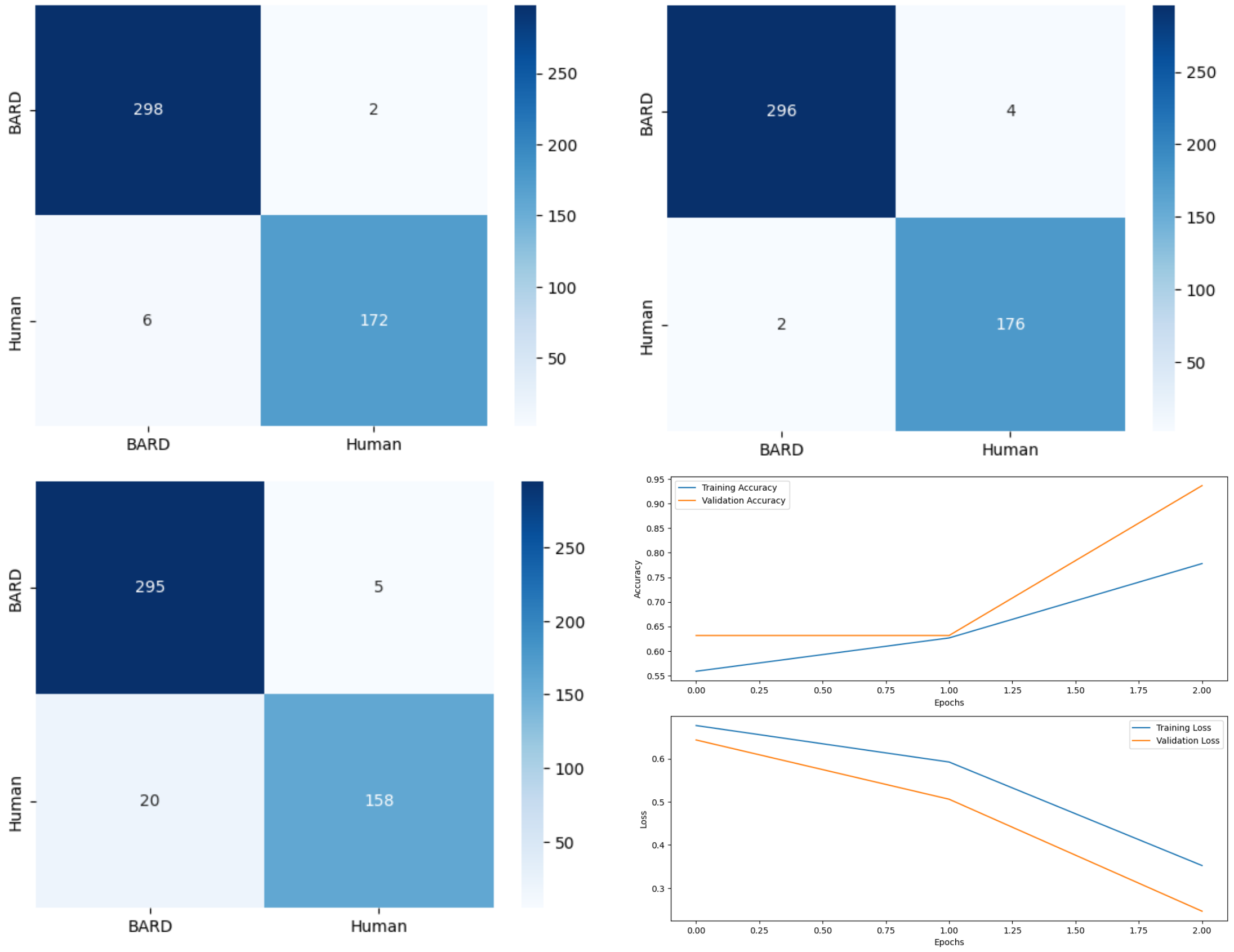

**Figure 27 Confusion Matrix for RF (top left), LR (top right), LSTM (bottom left), and Training Curve for LSTM (bottom right) on binary story classification (Human vs BARD)**

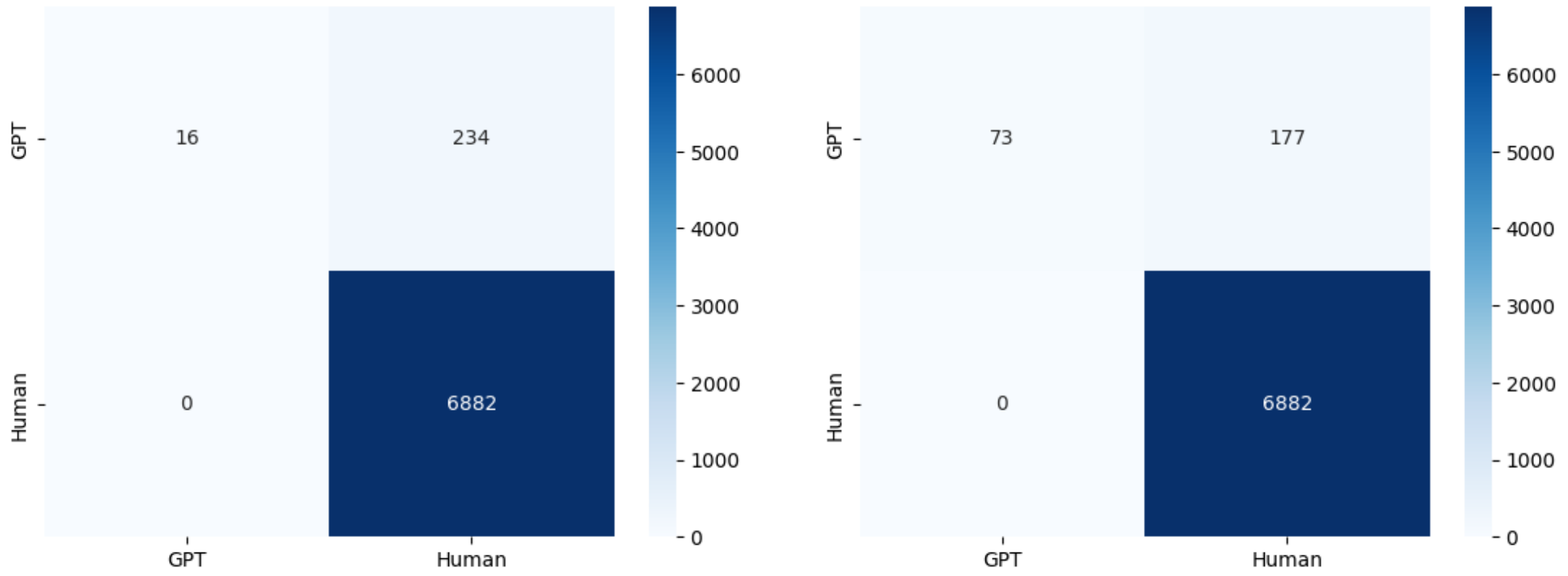

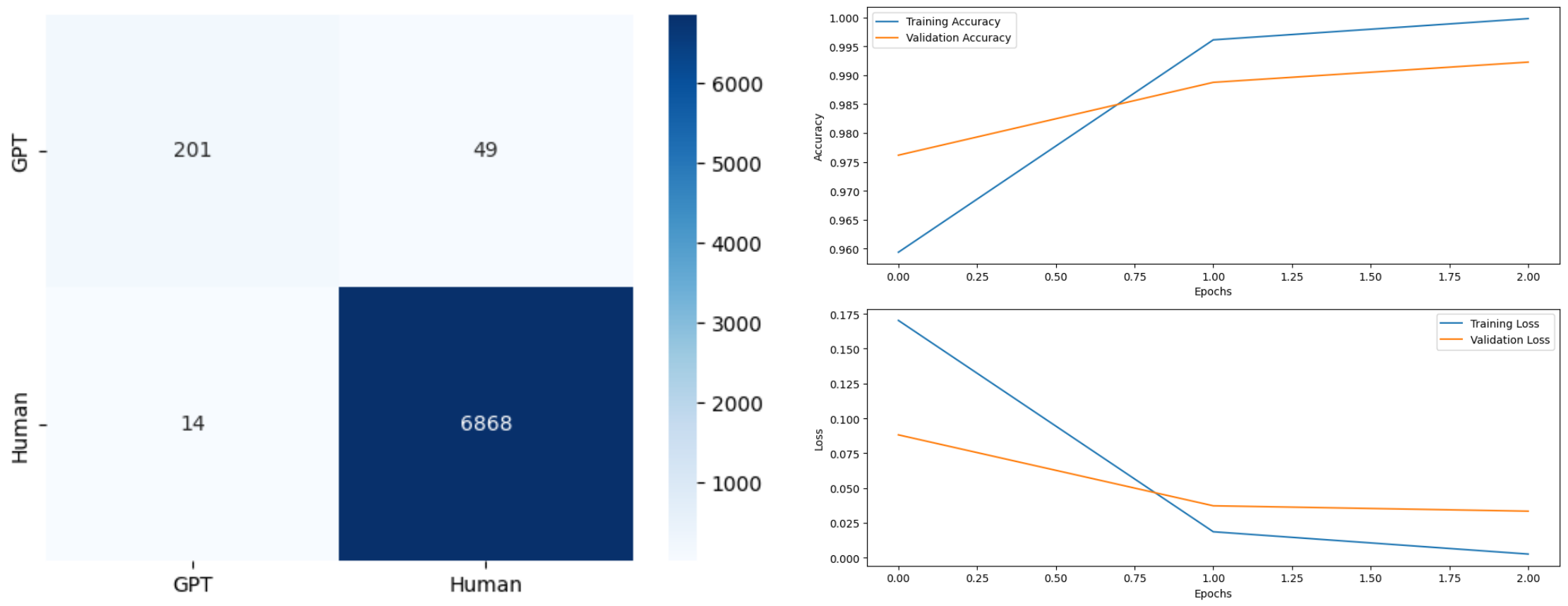

**Figure 28 Confusion Matrix for RF (top left), LR (top right), LSTM (bottom left), and Training Curve for LSTM (bottom right) on binary poem classification (Human vs GPT)**

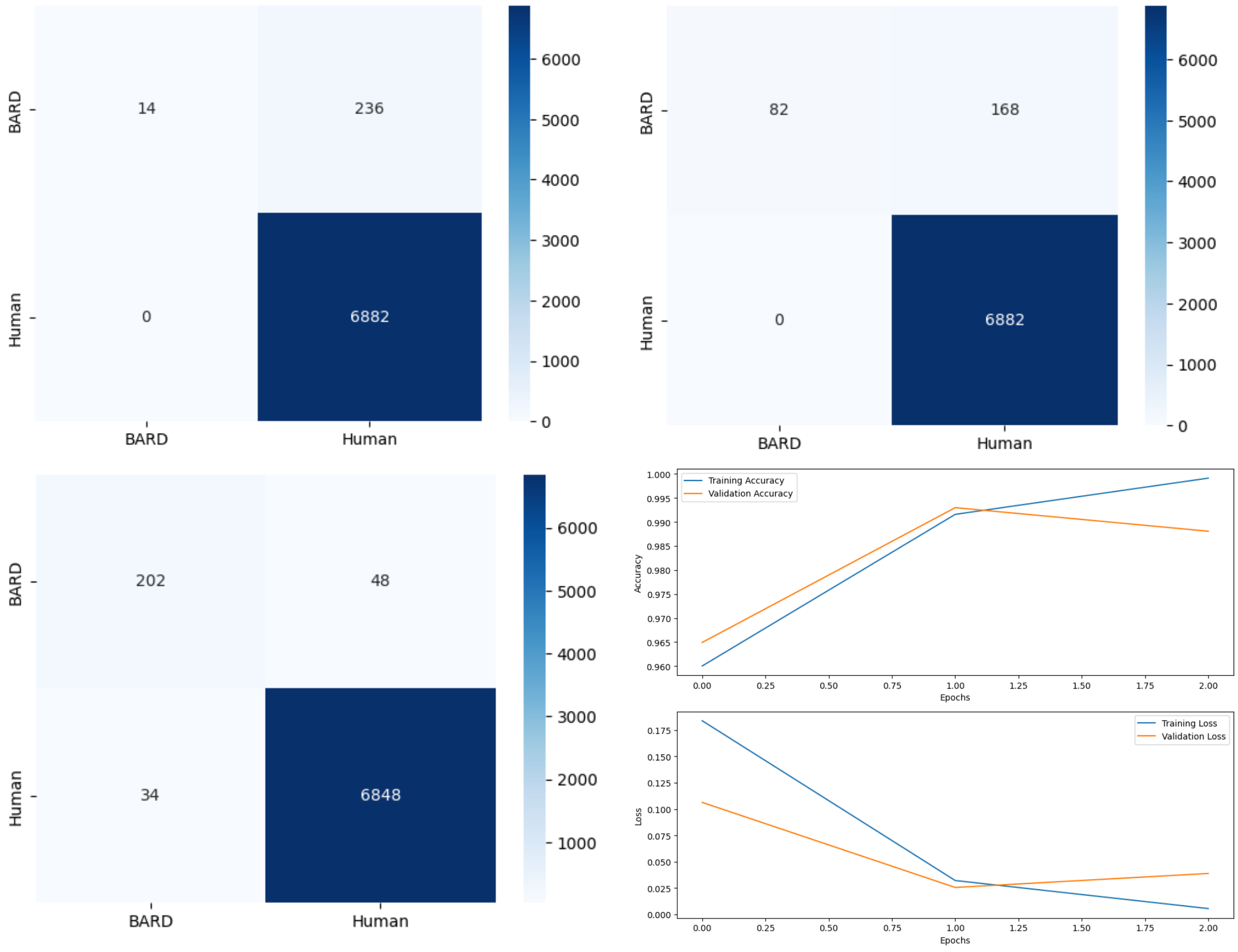

**Figure 29 Confusion Matrix for RF (top left), LR (top right), LSTM (bottom left), and Training Curve for LSTM (bottom right) on binary poem classification (Human vs BARD)**